\documentclass[final,1p,times]{elsarticle}

\usepackage[T1]{fontenc}
\usepackage{graphicx}
\usepackage{caption}
\usepackage{subcaption}
\usepackage{tabularx}
\usepackage{algorithm,algpseudocode}
\usepackage{url}
\usepackage{color}
\usepackage{times}
\usepackage{amsmath}

\definecolor{rulecolour}{rgb}{0.0,0,0.0}
\definecolor{headingcolour}{rgb}{0,0,0.0}

\newcommand{\assign}{\ensuremath{\leftarrow}}
\newcommand{\rlclr}[1]{\textcolor{rulecolour}{#1}}
\newcommand{\alg}[1]{\rlclr{\sffamily{#1}}}

\graphicspath{{./pix/},{../pix/}}

\usepackage{amssymb}

\journal{Knowledge-Based Systems}

\begin{document}

\begin{frontmatter}

\title{Local Universal Explainer (LUX) -- a rule-based explainer with factual, counterfactual and visual explanations}

\author[a]{Szymon Bobek}
\ead{szymon.bobek@uj.edu.pl}
\author[a]{Grzegorz J. Nalepa}
\ead{grzegorz.j.nalepa@uj.edu.pl}

\affiliation[a]{organization={Faculty of Physics, Astronomy and Applied Computer Science, Institute of Applied Computer Science, and Jagiellonian Human-Centered AI Lab, and Mark Kac Center for Complex Systems Research},
            addressline={ul. prof. Stanisława Łojasiewicza 11}, 
            city={Krakow},
            postcode={30-348}, 
            country={Poland}}

\begin{abstract}
Explainable artificial intelligence (XAI) is one of the most intensively developed area of AI in recent years.
It is also one of the most fragmented with multiple methods that focus on different aspects of explanations.
This makes difficult to obtain the full spectrum of explanation at once in a compact and consistent way.
To address this issue, we present Local Universal Explainer (LUX), which is a rule-based explainer that can generate factual, counterfactual and visual explanations.
It is based on a modified version of decision tree algorithms that allows for oblique splits and integration with feature importance XAI methods such as SHAP.
It limits the use data generation in opposite to other algorithms, but is focused on selecting local concepts in a form of high-density clusters of real data that have the highest impact on forming the decision boundary of the explained model and generating artificial samples with novel SHAP-guided sampling algorithm.
We tested our method on real and synthetic datasets and compared it with state-of-the-art rule-based explainers such as LORE, EXPLAN and Anchor.
Our method outperforms the existing approaches in terms of simplicity, fidelity, representativeness, and consistency.

\end{abstract}


\end{frontmatter}

\section{Introduction}
\label{sec:intro}

In recent years, a vast amount of research has been devoted to the development of explainable artificial intelligence (XAI) methods.
Their goal is to bring transparency and interpretability into the decision-making process governed by black-box machine learning models.
This resulted in high expectations for applicability of XAI-based system to high-risk areas such as industry 4.0, medicine, financial markets, etc.
These expectations were inflated not only by the rapid development of XAI algorithms, but also by legal regulations such as GDPR regulations~\cite{goodman2016regulations}, or the more recent AI ACT~\cite{aiact2022hacker}.

However, almost immediately the high hopes and expectations of making the ML/AI systems understandable for humans were cooled down when multiple empirical observations and expert-based critical reviews showed that the explanations very often cause more confusion than clarification. 
This has been noted by communities in almost each of the high-risk areas~\cite{ghassemi2021xaimedpitfalls,verma2021industryxaipitfalls,roski2021xaimedfail,evans2022xaifailpathology} but also by the creators of XAI systems~\cite{ehsan2021xaipitfalls,molnar2022pitfalls}.

One of the reasons for this situation is that the state-of-the-art methods focus only on one particular aspect of explainability, i.e. feature importance, rule-based explanation, counterfactual explanations, etc., and to obtain the full spectrum of explanations for a decision of a model, one has to combine the results of multiple different algorithms.
However, they are usually based on different theoretical backgrounds and often use statistical learning to produce explanations, making them inconsistent (or sometimes contradictory) with each other.
Additionally, all available local explainers use artificially generated samples to create explanations.
This may cause the locality-fidelity paradox, as recently reported in~\cite{gaudel2022paradox}, where local explanations incorrectly explain the behavior of a model, are overcomplicated and counterintuitive.
It also causes the explanations to be unstable, as the randomness related to data generation may affect the way the explanation is created for the same instance in consecutive runs of the explanation mechanism.
This penomenon is often referred to as Rashomon effect~\cite{rashomon2024mueller}.

Such properitis of state-of-the-art explanations limit their practical value becasue in real-case scenarios the explanation needs to be multifaced and cover different aspect of the decision of the balckbox model.
This challenge was previously noticed by different groups of researchers, but their focus was mainly on the integration of different explantions into a comprehensive dashboards~\cite{wang2019xaihci}, or on the interactive pipelines~\cite{baniecki2023grammar}.
However, such integration combine explanations that may yield different or contradictory results dut to their differences in design and the aforementioned Rashomon effect.
Therefore, in this work, we address these issues by proposing an universal explanation model that unifies most common explanation mechanism into one model enforcing their consistency.
The method is model-agnostic, which makes it possible ot aplpy it to any balckbox model and generates simple, rule-based explanations that are local, counterfactual, and consistent with SHAP explainer.

\begin{figure}
    \centering
        \includegraphics[width=1.0\columnwidth]{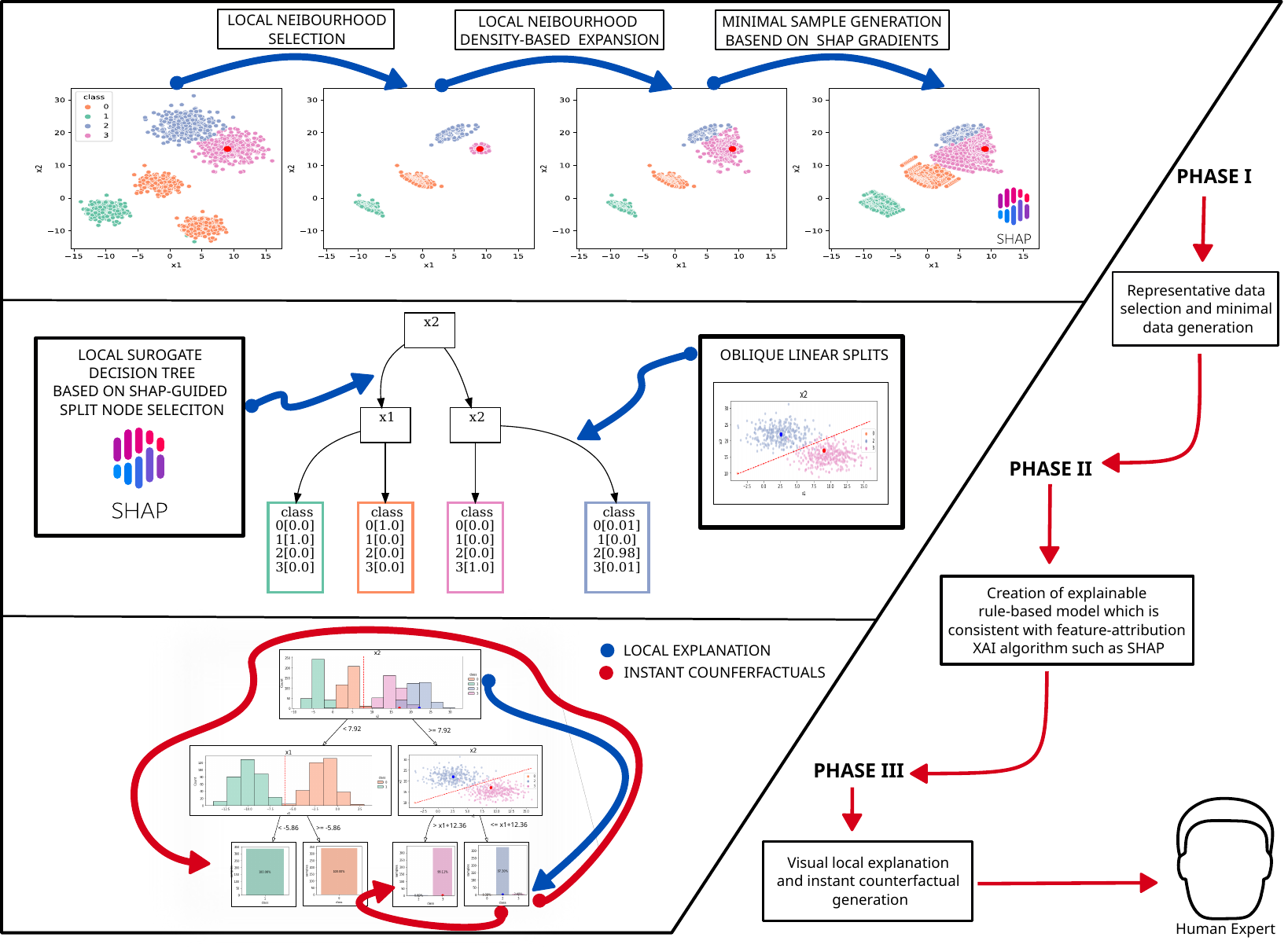}
\caption{Diagram representing three phases of rule-based explainable model creation. The first phase is focused on selecting and refining representative dataset for local surrogate model creation. The second phase concentrates on generating an oblique, shap-consistent decision tree. The third phase provides visual presentation of local explanations as well as counterfactual generation mechanisms.}\label{fig:overview}
\end{figure}

The overview of the contribution of this paper is presented in Figure~\ref{fig:overview}. 
The overall procedure is divided into three phases starting from data selection and preparation, through explainable local surrogate model training and visual presentation of explanations and counterfactuals.

As a backbone of our explanation mechanism, we used a decision tree structure.
It allowed us not only to generate rule-based explanations, but also to immediately extract consistent counterfactuals by traversing the explanation tree in an appropriate way.
We limited the use generated or perturbed data for training of the explanation model to avoid disrupting the distribution of the data and obtain the most plausible and representative counterfactuals.
We compensated for the possible shortage in training instances by using SHAP importance values as a support in selecting features to construct explanation rules and generating samples only towards the direction of negative gradient of shap values which points towards the direction of a decision boundary of a balckbox model.
This allowed us to obtain explanations consistent with SHAP, which is one of the most popular methods used in the XAI field.
Finally, we extended the possible tree nodes to contain simple oblique splits (based on linear inequalities rather than value-based inequalities), which we visualize on a 2D scatterplots.
We show that our approach outperforms other state-of-the-art rule-based solutions with respect to global fidelity, representativeness, and simplicity and consistency with SHAP. 
We present the results obtained from a publicly available, fully reproducible benchmark.
To the best of our knowledge, this is the first comprehensive comparison of local rule-based explainers available in the literature so far, not counting the work of G. Vilone, et. al~\cite{Vilone2021AQE,vilone2022hceval}, which focuses on global rule-based explanations in contrast to our contribution.

The remainder of the paper is organized as follows. 
In Section~\ref{sec:sota} we describe the wide spectrum of existing XAI methods and place our work in that context.
In Section~\ref{sec:motivation} we provide a critical overview of rule-based explainers and form four requirements that every local, rule-based explainer should meet. 
We present our method in Section~\ref{sec:lux} followed by the evaluation given in Section~\ref{sec:eval}.
We summarize our work by providing its strong and weak sides in Section~\ref{sec:summary}.

\section{Related works}
\label{sec:sota}

The primary goal of any explainable AI method is to provide some sort of insight that can be used by a human to understand how the model processes input data to the output.
This insight can have a very different form and can carry different amounts of information depending on the XAI algorithm used.
It can also answer very different questions about the operational aspects of the AI model at very different levels of granularity.
In this section we scetch the landscape of XAI methods, pointing out the most knwon algorithms and explanation types.
However, this description does not not fully exhaust this topic, and we refer interested reader to~\cite{barredo2020infus} or~\cite{gonzales2019bbwb} that provide comprehensive review of existing approaches and trends in state-of-the-art XAI.

From the perspective of granularity of explanation, one can classify the XAI algorithms into local explanations that focus on generating explanation around a particular instance and global explanations that aim at providing general explanation for a model as a whole. The most representative examples of the former are SHAP~\cite{shap} or LIME~\cite{lime}, while the latter are mostly dominated by well established statistical methods such as partial dependence plots, accumulated local effects, functional decomposition, importance of permutation features or prototype discovery~\cite{molnar2020interpretable}.

From the perspective of the universality of the explanations, one can distinguish model-agnostic algorithms that do not depend on the model to which they can be applied and model-specific methods such as those which are designed to work with a particular type of the model.
Model-agnostic methods treat the model they explain as a balck-box, without any assumptions on its internal mechanisms.
This makes them very universal solutions as they can be applied to any kind of system that converts a predefined set of inputs into the output. 
Examples of model-agnostic methods are algorithms already mentioned,  such as SHAP or LIME,  Anchor~\cite{ribeiro2018anchors}, LORE~\cite{guidotti2018lore},  EXPLAN~\cite{rasouli2020explan} but also most global explainers~\cite{ribeiro2016modelagnostic}.
However, model-specific methods leverage some of the properties of the algorithm they explain, which can boost their performance. 
In some cases, the model-agnostic method can be redesigned to be model-specific.
The example of this is the SHAP method which has a model agnostic version in the form of KernelSHAP~\cite{kernelshap} and model-specific version such as TreeSHAP~\cite{treeshap}, or DeepSHAP~\cite{shap} that are designed to work faster with different families of ML algorithms.

This classification can be further broken down into inherently interpretable models and post hoc explainers.
The first group is formed by the AI systems that are interpretable by design, which allows one to explain its decision by tracking the process they were made.
Linear regression and decision trees are the most commonly used methods among them.
The trend of building efficient glass-box, inherently interpretable models instead of post-hot approaches is growing in popularity.
This was first tackled in~\cite{rudin2019explaining}, where the author stated that there is no scientifically proven dependency between model complexity and its efficiency. 
A similar approach was presented in~\cite{caruana2015ebm}, where the authors present explainable boosting machines, a method that is interpretable and efficient.
The post-host model assumes that the explanation can be generated without any interference in the internal mechanics of the model that is being explained.
However, the existence of post-hot and inherently interpretable models in one explanation mechanism is not mutually exclusive.
In fact, a building block of many post-hoc eplainers, such as model-agnostic explainers mentioned before, are inherently interpretable models which locally mimic the behavior of the original model.

Finally, from the perspective of the form of the explanation one can distinguish three leading strategies of generating and presenting explanations: 1) feature importance attribution methods, 2) counterfactual and contrastive explanations, and 3) rule-based explanations.
In the first group, the leading places are taken by algorithms such as SHAP, LIME or BreakDown~\cite{biecek2019breakdown}.
Although they differ in the underlying mechanisms, they all provide local explanations in the form of feature importances and feature impacts.
Such explanations can be used to answer the question of how different features affect the model decision for a particular instance. 
Although this form of explanation may seem to be one of the simplest, they are also one of the most popular ones.
However, this simplicity is illusory and is prone to many serious interpretation errors, which was noticed in many recent works both from the data-science perspective~\cite{molnar2022pitfalls,ehsan2021xaipitfalls} as well as from the perspective of domain experts from different fields such as industry~\cite{verma2021industryxaipitfalls} or  healthcare~\cite{ghassemi2021xaimedpitfalls}.

The group of counterfactual and contrastive explanations is formed by approaches such as~\cite{stepin2021cfxaisurv}.
The main goal of this type of explanation is to provide information on how to change the features of a given instance to change the classification result of the model. 
Among the most popular counterfactual explainers, where are: DiCE (Diverse Counterfactual Explanations)~\cite{dice2020}, which provides an explanation in the form of minimal perturbations needed to change the decision of a model for a particular instance; Actionable Recourse in Linear Classification~\cite{action2019cf}, which provides a method of generating counterfactuals only perturbing the \emph{actionable} variables that can be changed in reality. 
More methods can be offered by CARLA (Counterfactual And Recourse Library)~\cite{pawelczyk2021carla} which is a framework that integrates multiple different methods for counterfactual explanations.
All of the aforementioned methods focus solely on the counterfactual explanations, not providing any other means of insight into the model decisions.

Finally, rule-based explainers are dominated by three state-of-the-art algorithms, already mentioned: Anchor, LORE and EXPLAN.
They provide explanations in a form of rules which are human-readable and can be interpreted easily even with non-data-scientists.
The Anchor algorithm uses a modified beam search mechanism combined with a pure-exploration Multi-Armed-Bandit~\cite{kaufmann2013mab} to find rules that describe the class of an instance that is explained in its local neighborhood.
It provides high-precision rules, which explain the decision of a black-box model for a particular instance; however, it does not provide counterfactual explanation at all.
LORE and EXPLAN algorithms use a decision tree as the backbone of the rules-building mechanisms.
This allows these methods to extract counterfactuals from the tree structure, not only the explanation of a current instance.
However, they both use data-generation approaches that change the distribution of the original data.
This may result in the appearance of false counterfactuals in the form of phantom branches in a tree -- branches that were created based solely on generated instances and have no coverage in the real dataset.
Furthermore, they tend to generate rules which are overcomplicated and not consistent with SHAP values calculated for the explained instance, which is caused by the greedy nature of the decision tree mechanism.
An interesting approach was presented in the RuleXAI~\cite{rulexai2022sikora} toolkit, where the authors used information on the relevance of features to filter attributes that should be used to form an explanation rule, which can improve consistency with methods based on the importance of features.
However, the authors did not consider that the importance of the features used to generate a local explanation should be considered only in the locality of the instance being explained, otherwise it may cause serious misinterpretation of the results as reported in~\cite{molnar2022pitfalls} and ~\cite{biecek2019breakdown}.
Furthermore, they used feature importance not corresponding to the type of the model they explain assuming that the important feature is utilized similarly by a non-parametric model like a decision tree and parametric one like neural network, or logistic regression, which is not true.
Finally, none of the above methods provides visualization of the explanations that can help in the interpretation of the data.
The synthsized comparison of the state of the art methods that our method competes with is given in Table~\ref{tab:comp}.

\begin{table}[]

\caption{Comparison of capabilities offered by state-of-the art implementations of rule-based XAI methods.}
\label{tab:comp}
\centering
\begin{tabular}{|c|c|c|c|c|}
\hline
                & \textbf{Factual} & \textbf{Counterfactual} & \textbf{Visual} & \textbf{Example-based} \\ \hline
\textbf{LORE}   & \textbf{Yes}               & \textbf{Yes}                      & No              & No                     \\ \hline
\textbf{EXPLAN} & \textbf{Yes}               & No                      & No              & No                     \\ \hline
\textbf{Anchor} & \textbf{Yes}               & No                      & No              & \textbf{Yes}                     \\ \hline
\textbf{LUX}    & \textbf{Yes}              & \textbf{Yes}                      & \textbf{Yes}              & \textbf{Yes}                     \\ \hline
\end{tabular}

\end{table}

In this work, we focus on post-host, model-agnostic, and local explainers.
Motivation behind that arises from the target users of the explanation mechanism we aim at.
Our goal was to design a method that will be addressed to non-data-scientist users, experts in the domain of interest, and that will solve the issues with existing rule-based explainers.
For this purpose, we propose a method that is 
model-independent, and highly human-readable and combines all forms of explanations into one consistent mechanism enhanced with visualization.
The following section will state the problem formally and confront it with the state-of-the-art approaches.

\section{Problem statement}
\label{sec:motivation}

In our work, we focus on explanations that can be used by non-technical user, particularly domain expert in a field.
This limited our research to rule-based explainers, as their ability to be understood by experts is one of the highest~\cite{buts2022understandability}, especially in the areas related to medicine, industry, and law where rule-based knowledge is used on a daily basis.

\subsection{Critical overview of the existing rule-based explainers}
\begin{figure}
    \centering
        \includegraphics[width=1.0\columnwidth]{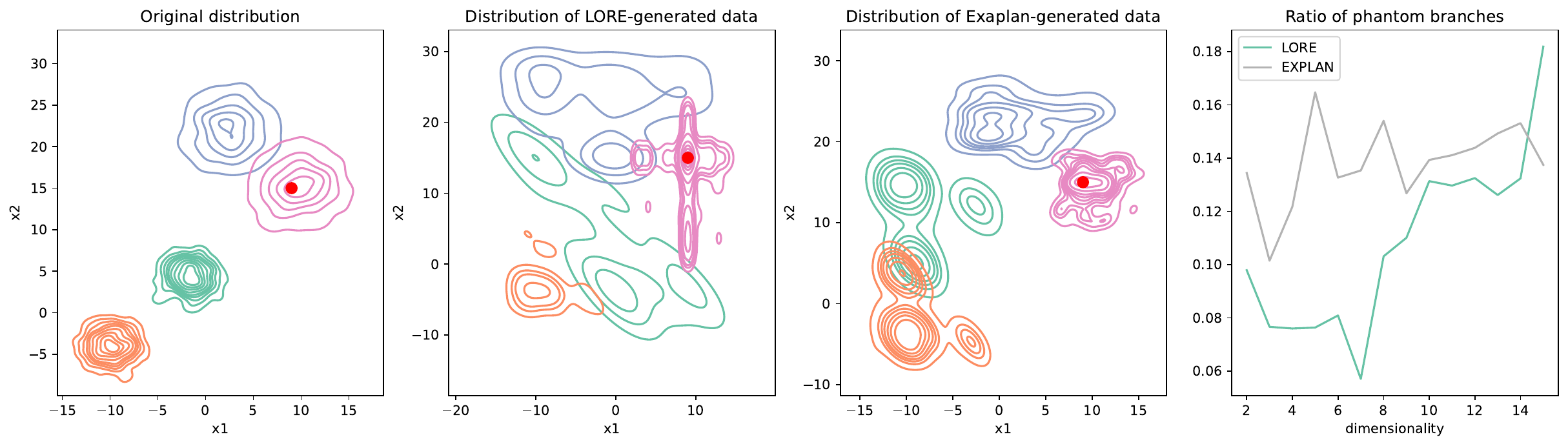}

\caption{Impact of the generated data on the explanations generated with LORE and EXPLAN. The leftmost plot represents distribution of the original data, the following two plots show distribution changed by the data-generation algorithm used by LORE and EXPLAN. The rightmost plot shows how many phantom branches are generated from such an altered distribution. Such  phantom branches may generate non-representative counterfactuals, like these marked in red circles.}\label{fig:degenerated-dists}
\end{figure}

In our work, we focus on local explanations, which are designed to explain a decision for a particular instance, as opposed to global explanations, whose objective is to explain an average behavior of the model on a given dataset.
One of the basic criteria of the correctness of an explanation is its faithfulness.
The explanation is considered faithful if $\Phi^{\overline{E}\rightarrow M(x_i)} = M(x_i)$, which means that the explanation provides the same prediction for an instance as the original model.
Therefore, it is assumed that local extension is more faithful in the nearest neighborhood of the instance that is being explained than in the whole dataset, because it is assumed that the decision boundary at a subset of the data can be approximated by a simple model, such as a linear model in case of LIME or rule-based models in case of LORE, EXPLAN or ANCHOR.
However, incorrectly selected neighborhood with many artificially generated samples may cause the production of an explanation model that is degenerated~\cite{gaudel2022paradox} causing the explanation to be unfaithful, counterintuitive and overcomplicated. 
Figure~\ref{fig:degenerated-dists} presents the results of the experiment we conducted with EXPLAN and LORE to observe how the generation of artificial samples may affect the explanations. It can be seen that an altered distribution of the data causes the generation of multiple phantom branches that do not have coverage in the real data.
In case of counterfactual explanation, this may result in building explanations that are not consistent with dataset and, hence, counterintuitive to the expert.
It also makes the whole decision tree larger and more complex.

\begin{figure}
  \centering
      \includegraphics[width=1.0\columnwidth]{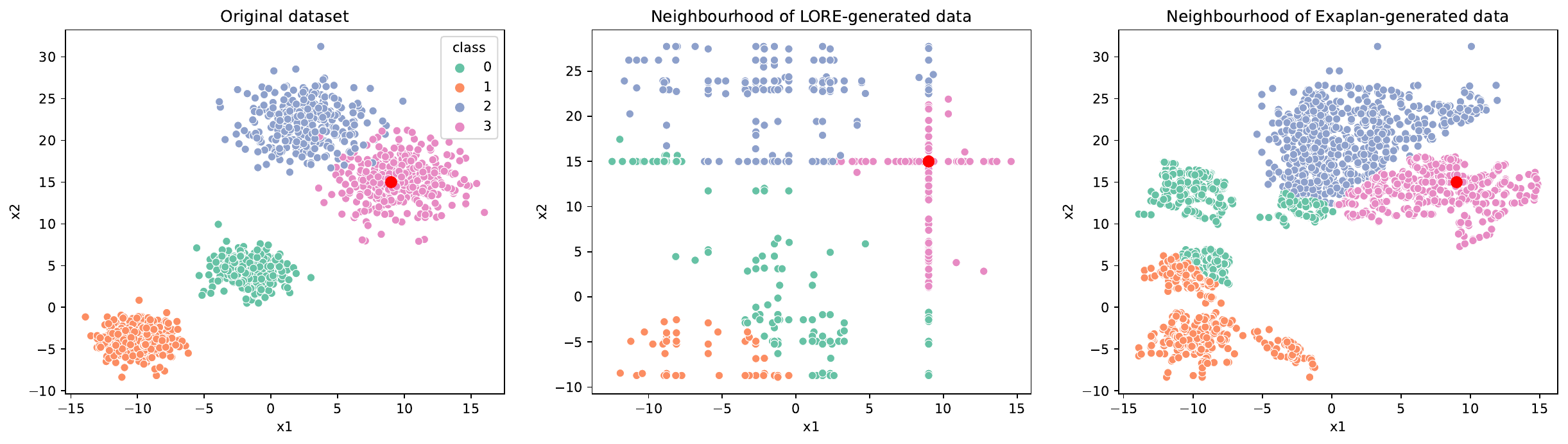}
\caption{The differences in the local neighborhood defined by different rule-based explainers. The local neighborhood was generated for the instance marked red. }\label{fig:degenerated-neigh}
\end{figure}

Moreover, the selection of the neighborhood for building local explanation cannot be simplified to just distance-based selection. The EXPLAN algorithm approaches this problem by including agglomerative clustering as a method of selecting representative samples. 
However, this is not always a correct strategy, as depicted in Figure~\ref{fig:degenerated-neigh}.
The figure shows that the neighborhood that is sufficient to locally explain the decision of the classifier can be limited to explaining the decision boundary between clusters 2 and 3. However, both LORE and EXPLAN will always include in the neighborhood other, possibly distant clusters of data, which do not necessarily should be considered as a local neighborhood.

\begin{figure}
  \centering
      \includegraphics[width=1.0\columnwidth]{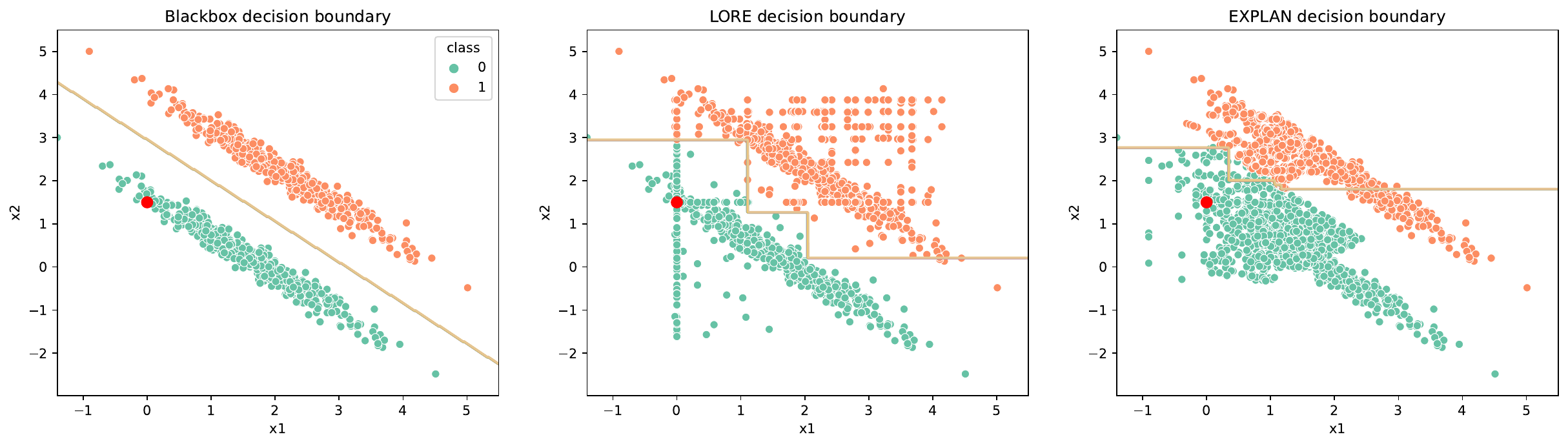}
\caption{Decision boundaries of a blackbox model and local approximations of that boundary with LORE, EXPLAN and Anchor. It can be seen that in case of linear boundaries, that are hard to capture by tree-based inequalities the existing explainers produce overcomplicated and non-intuitive explanations. }\label{fig:degenerated-decbounds}
\end{figure}

Furthermore, all of the rule-based explanation mechanisms limit the presentation of the explanation to a form of textual rule.
In most of the practical applications, such a textual representation might not be enough, especially when working with continuous values, where numerical thresholds set on rules' conditions are not always meaningful to the experts.
The explanation has to be understandable to the user in order to be useful and to have practical value.
Focusing only on one aspect of the quality of explanation can lead to degenerated results, as was shown in~\cite{aytekin2022neural}, where the authors represented a neural network as a decision tree with split conditions represented as linear algebra inequalities.
Although the method achieved good results in terms of fidelity, from a practical perspective, it was yet another black-box model. 
Understandability is hard to measure; however, as reported in several empirical studies~\cite{collaris2020explexpl,collaris2022comparative,buts2022understandability} the most understandable explanations are those that are visual, contrastive and rul-based.
Despite that fact, none of the state-of-the art rule-based explainer includes visual representation of explanation. 
Additionally, the rules are generated by decision tree algorithms, or rule-discovery algorithms that generate explanations in the form of conjunction of simple inequalities.
This feature is motivated by the simplicity argument that states that the rule-based format of explanations is one of the most understandable~\cite{buts2022understandability}.
However, in many cases, this is not true, especially when the approximated decision boundary is a liner function.
In such a case, the rule-based approximation may overcomplicate it rather than simplify. 
The example of such a case was given in Figure~\ref{fig:degenerated-decbounds} where the LORE, EXPLAN and Anchor decision boundaries were shown along with the original decision boundary of the model they explain.

\begin{figure}
  \centering
      \includegraphics[width=1.0\columnwidth]{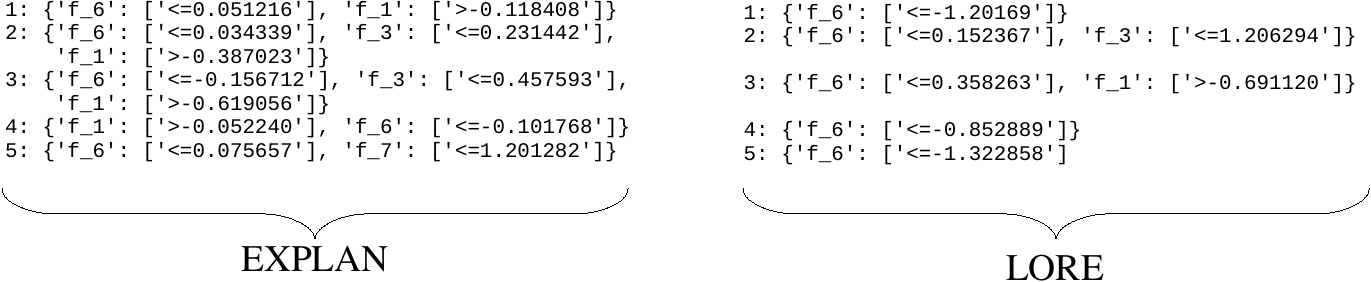}
\caption{Five explanations generated for the same instance and the same dataset. It can be seen that the explanations are different in consecutive runs, making them less usable and less intuitive for users.}\label{fig:lore-explan-consistency}
\end{figure}

Another aspect that is observable with state-of-the-art explanation mechanisms is the consistency issue. 
It is defined as similarity of the explanations generated by the explainer for the same instance (e.g. is the explanation for a given instance the same for consecutive calls of the explainer). 
From our experiments, we discovered that not only the state-of-the-art explainers are not consistent between each other, but due to the randomness of the data-generation technique they use, the explanations form the same explainer might be different for the same instance and the same initial dataset. 
This situation was illustrated in Figure~\ref{fig:lore-explan-consistency}. 

In addition, the explainers focus mostly on one type of explanation: the factual one.
Only LORE provides counterfactual explanations, while EXPLAN and Anchor do not.
However, the counterfactual explanations of LORE suffer from the same problems as the factual explanation.

In the following section, we define several requirements for the rule-based explanation to meet in order to overcome the limitations discussed above.

\subsection{Requirements definition}
\label{sec:measures}
In our case, the explanation $E$ for model $M$ that was trained on dataset consisting of features  $F \in \{f_1, f_2, \ldots, f_n\}$ for an instance $x_i$ is considered to be of the following form given in Equation~(\ref{eq:explanation}).

\begin{equation}
\label{eq:explanation}
\Phi^{E\rightarrow M(x_i)}: C(F_1) \wedge C(F_2) \wedge \ldots \wedge C(F_k) \rightarrow l
\end{equation}

\noindent 
Where $C(F_i)$ is a condition in a form of inequality that includes a subset of features $F_i \in F$. In the simple case, $F_i$ will contain only one of the features from $F$, and $l \in L$ is one of the class labels used by model $M$.

In the following paragraphs, we define the most important requirements that explanations should meet.
We focus on the requirements in the context of rule-based explanations.
In our work, we use the definitions of XAI metrics from~\cite{coroma2022xaimetrics} that we adopted to the needs of a rule-based format.

\paragraph{R1: Simplicity}
The explanations must be simple in order to be understandable and therefore useful to the expert. 
The rule-based explanation $\Phi_1$ is considered simpler when $\Phi_2$ if the number of features used to represent $\Phi_1$ is less than the number of features used to represent $\Phi_2$.
Therefore, for $\Phi_1^{E\rightarrow M(x_i)}: C(F_a) \wedge C(F_b) \rightarrow l$ and $\Phi_2^{E\rightarrow M(x_i)}: C(F_i) \wedge C(F_j) \rightarrow l$ the $\Phi_1$ is simpler when  $\Phi_2$ if $\left \| F_a \cap F_b \right \| < \left \| F_i \cap F_j \right \| $.
Simple explanations are more comprehensible for users and therefore more useful.

\paragraph{R2: Counterfactual and representative} 
The explanation must provide a counterfactual explanation, preferably in the form of an example from the original dataset. 
The counterfactuals explanation $\overline{E}$ for model $M$ that was trained on dataset consisting of features  $F \in \{f_1, f_2, \ldots, f_n\}$ for an instance $x_i$ is considered as $\overline{x_i}$ which satisfies the rule given in Equation(\ref{eq:counterfactual}).

\begin{equation}
\label{eq:counterfactual}
\Phi_{CF}^{\overline{E}\rightarrow M(x_i)}: C(F_1) \wedge C(F_2) \wedge \ldots \wedge C(F_k) \rightarrow \neg l
\end{equation}

Where $C(F_i)$ is a condition in a form of inequality that includes a subset of features $F_i \in F$ and $\neg l \in L$ is one of the class labels used by the model $M$, but different than $l$.

The counterfactual explanation should also be representative, meaning that it should provide a representative example (or examples) from the dataset, otherwise it might be counterintuitive for the expert.
This may happen when the counterfactual explanation covers an input space which is populated only with artificial samples, which existence is not possible in real world, due to some constraints that are not captured by the model (like physical constraints).

\paragraph{R3: Consistency}
The explanation is considered consistent with other explanation methods if it uses the same set of attributes in the same way. 
Therefore, for two rule-based explanations $\Phi_1^{E\rightarrow M(x_i)}: C(F_a) \wedge C(F_b) \rightarrow l$ and $\Phi_2^{E\rightarrow M(x_i)}: C(F_i) \wedge C(F_j) \rightarrow k$, the explanations are consistent with each other if $k = l $ and $\left \| (F_a \cap F_b) \cup (F_i \cap F_j) \right \|$ is maximal and $J(D_1,D_2) = \frac{\left \| D_1 \cap D_2 \right \|}{\left \| D_1 \cup D_2 \right \|}$ is maximal, where $D_i$ is a subset of dataset covered by the rule defined by $\Phi_i$ and $J(D_i,D_j)$ is a Jaccard index for sets $D_i$ and $D_j$.
In other words, the two rule-based explanations are consistent with each other if they use the same set of attributes to describe the same subset of data.

\paragraph{R4: Visual representation}
The explanation in addition to the quality assessment metrics must be understandable for the addresee.
In our previous works~\cite{bobek2022knac,sbk2022clamp,jjb2022anodet}, we encountered this issue several times when working with domain experts. 
Even with the extensive explanations provided by the most popular XAI algorithms, they had difficulty in spotting the real mechanics that drive the model towards certain decisions.
Therefore, presenting explanations at the semantic level, aligned with the conceptual profile of the addressee of the explanations (the human), is crucial for improving understandability of XAI models for non-technical users.  
As mentioned previously,  understandability can be improved by choosing the representation of explanations that  improves the ability of the users to perceive the
knowledge gained by a model~\cite{paez2019understandability} such as a rule-based explanation~\cite{buts2022understandability}.
Another direction of research involves studies related to the application of visual explanatory methods.
Most of the works in this area focus on the visual presentation of such results, which has been proven to be one of the most effective channels for communicating knowledge in the field of XAI and ML~\cite{collaris2020explexpl,collaris2022comparative}. 
Work on visualization of explanations includes saliency maps for deep neural networks~\cite{gradcam}, task-specific visualizations~\cite{collaris2020explexpl,suh2019anchorviz}, as well as the various plots offered by SHAP and LIME.

In the next Section, we introduce Local Unvertain Explaination (LUX) mechanism -- a novel method for generating rule-based factual and counterfactual explanations that addresses all of the aforementioned requirements and issues.

\section{Local Universal Explanations}
\label{sec:lux}
In order to solve the issues of the existing state-of-the-art approaches and fulfill the requirements R1-R4, we propose an algorithm consisting of three phases.

\paragraph{Phase 1}
Representative data selection and minimal data generation. Our goal was to explain the decision of the model using the majority of the original data. Therefore, we focused on identifying a representative neighborhood of the instance that is being explained. The generation of samples is reduced to the minimum. This contributes to the simplicity and consistency of an explanation.

\paragraph{Phase 2}
Creation of an explainable rule-based model that is consistent with ferature-attribution XAI algorithms (e.g. SHAP or LIME),  provides consistent, example-based counterfactuals, and includes uncertainty of the black-box model in both factual and counterfactual explanations. This contributes to simplicity, consistency, and quality assessment requirements.

\paragraph{Phase 3}
Explanation visualization. We transferred the weight of the explanation from textual form to visual form. In particular, the rules we generate can contain two-dimensional linear inequalities, which are hardly comprehensible in text format but difficult to grasp in visual form. This contributes to simplicity and visualization requirements. 

\subsection{Representative data selection}
The problem of selection of representative sample of data to train a local explainer is non trivial task.
As discussed before and shown in Figure~\ref{fig:degenerated-dists}, the state-of-the-art rule-based explainers use artificially generated data as a base for selecting representative sample for training, which corrupts the distribution of the data and makes it difficult to identify the correct neighborhood for local explanations.
This may result in overcomplicated explanations and non-truly local explanations, which could be valuable in debugging a ML model, but less informative for domain experts who are mostly interested in what concepts (patterns) in the data are used by the model to provide a certain decision.
This question cannot be answered with artificially generated data that domain experts are not familiar with or whose existence in the real world is disputable.
Furthermore, artificially generated samples results in large ratio of so called phantom branches, i.e. branches that have no coverage in real exmaples and was genreated only based on artifically populated data.
The example of such a situation was given in Figure~\ref{fig:degenerated-dists}, in the right-most plot, where the ration of pahntom branches is presented with respect of number of dimensions of the dataset. 
It can be seen that artificially generated samples makes the ratio of such branches increasing for in LRE and EXPLAN possibly creatign explanations that have no real representatives in the dataset.

In our method, we wanted to build an explainable model based on concepts that are already there in the data.
We noticed that high-density areas of the input space can be considered as prototypes or concepts that can be used to explain the classification.
We used OPTICS~\cite{optics1999}, a density-based clustering algorithm that provides correct results if the distribution is the same as in the case of the original data.
Therefore, we inverted the process of selecting a sample for the explanation mechanism compared to the state-of-the-art approaches.
We first select the prototype concepts around the neighborhood of the instance that is being explained, based on the original data, and then, if necessary, we generate additional samples to balance the dataset or to resolve the uncertain decision boundaries. 
For data generation, we modified the BorderlineSMOTE~\cite{borderlinesmote2005} algorithm to up-sample only isolated (surrounded by the opposite class) and uncertain instances (instances for which the blackbox model prediction confidence is low).
In the following paragraphs, we describe this process in detail.

Let us assume that an instance for which a decision of a model is to be explained is defined as $x_i$, and the label of the instance predicted by a blackbox model $M(x_i) = l$.
We start by generating a base neighborhood $N$ of size $K$, which will become a seed for the density-based neighborhood creation method.
We define the base neighborhood $N$ of size $K$ as presented in the Equation~(\ref{eq:basneighbourhood}).

\begin{equation}
\label{eq:basneighbourhood}
N^b(x_i,K) = \left \{\forall_{c \in C},  x_k \in X : \epsilon \geq d(x_i,x_k) \leq D_{i,c}^{(\frac{K}{\|C\|})}  \right \}
\end{equation}

\begin{figure}
  \centering
      \includegraphics[width=1.0\columnwidth]{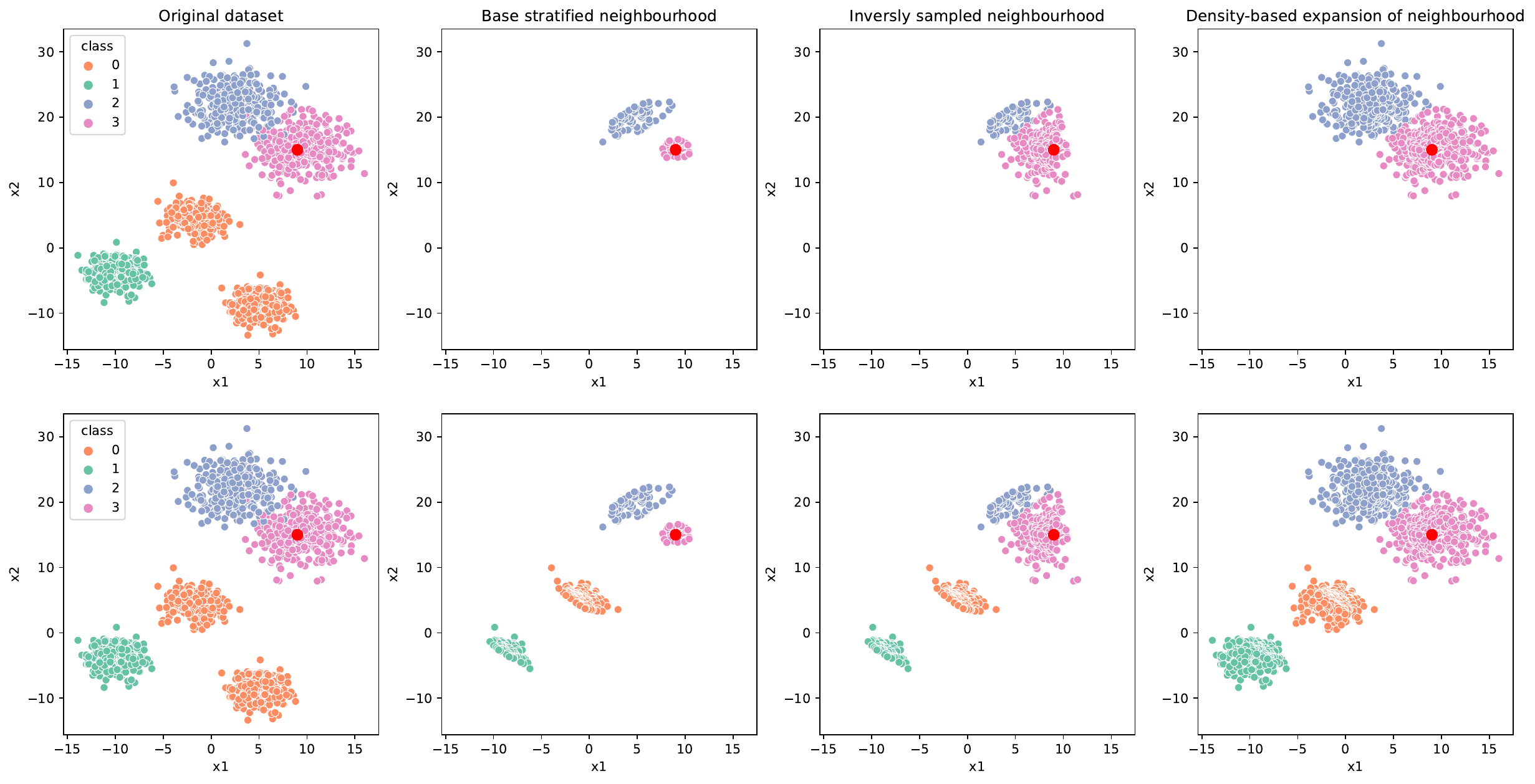}

\caption{Visualization of a neighborhood selection process in LUX. The upper row shows the selection of the neighborhood according to local stratification and the bottom row shows the process for global stratification. Consecutive plots show consecutive steps of selecting neighborhood.}\label{fig:neighbourhood-lux-data}
\end{figure}

\noindent
Where $D_{i,c}^{(l)}$ is $l$-th element from a tuple $D_{i,c}$ defined for all $m$ instances from a training set that were labeled as $c$ by a blackbox model $M$:
$$D_{i,c} = \left \{ d(x_i, x_1), d(x_i, x_2), \ldots, d(x_i, x_m) \right \}$$
\noindent
$D_{i,c}$ is sorted in ascending order, and $d(x_i, x_j)$ is a distance between instances $i$-th and $j$-th.
The base neighborhood $N^b(x_i,K)$ is created in a stratified way to ensure the existence of the desired class representatives.
The $\epsilon$ defines a stratification threshold and by default it is set to $\infty$, meaning that no constraints on the distance of the representative samples from the instance that is being explained are set.
The stratification can be local or global, as depicted in Figure~\ref{fig:neighbourhood-lux-data}. 
In the local stratification, the data is sampled in a range defined by a distance from the instance being explained to the nearest instance of the opposite class.
This stratification is desired in the case of simple local explanations and local counterfactuals.
In the global stratification, samples from all classes are sampled to become a part of the $D_{i,c}$ training set and the $\epsilon = \infty$ this stratification is more appropriate if counterfactuals should cover a whole spectrum of possible classes.

In addition to stratification, the inverse sampling strategy is also applied.
Inverse sampling allows us to better spot the decision boundaries between classes.
In this approach, $N^b(x_i,K)$ is extended by $N^{-1}(x_k,K)$ defined in the Equation~(\ref{eq:invneighbourhood}).
\begin{equation}
\label{eq:invneighbourhood}
N^{-1}(x_k,K)= \left \{ x_k \in X : M(x_i) = l,  d(x_k,x_i) \leq D_{i,l}^{(\frac{K}{\|C\|})}  \right \}
\end{equation}

\noindent
In other words, the neighborhood is extended with points that surround the decision boundary that divides the classes $M(x_i) = l$ and $M(x_k) = c$, where $l$ is the class of an instance being explained. The inverse sampling strategy is depicted in Figure~\ref{fig:neighbourhood-lux-data} in the third column of the plots.

Up to this point, the neighborhood was constructed solely on the basis of the distances between samples.
However, such an approach does not take into account the density of the data, which could affect the way the locality is perceived.
In particular, high-density areas that were not included in the distance-based neighborhood creation step may contain a significant mass of samples that influences the shape of the decision boundary.
This situation is presented in Figure~\ref{fig:neighbourhood-lux-data} in the last column of the plots.
To address this issue, we perform a density-based clustering on the entire training dataset, to obtain a set of $n$ high-density areas of instances $\left \{C_1, C_2, \ldots, C_n \right \}$.
The number of $n$ is governed by the parameter $\sigma$ which defines a minimal number of instances that can be used to form a cluster (standalone high-density area).
Then, we extend the base neighborhood to contain only the sets of these high-density areas that  already have some nonempty intersection with points from base neighborhood.
Therefore, the final neighborhood $N$ is now defined as:

\begin{equation}
\label{eq:fullneighbourhood}
N = N^b \cup N^{-1} \cup \left \{  C_{i = 1,2,\ldots,n}: C_i \cap \left \{ N^b \cup N^{-1}\right \} \right \} \neq \varnothing 
\end{equation}

Finally, the oversampling algorithm is used to balance the dataset and to generate samples in regions that are highly uncertain.
This method does not populate empty regions with artificially generated data like LORE or EXPLAN, but tries to upsample data within the limits defined by the neighborhood $N$.
We extended the BorderlineSMOTE algorithm, which originally generates synthetic samples only around samples called \emph{in-danger}.
The \emph{in-danger} sample was a sample whose K-nearest neighbors were from the opposite class.
We exploit the observation that the areas which require data generation are those located around the decision boundary. 
The excat location of decision boundary in multidimensional, non-linear case is non-trivial taks, however, its location can be approximated by the confidence of a prediction of the blackbox model.
If the prediction of a black-box model is uncertain for a sample $x_i$ it means that the sample is located somewhere in proximity of a decision boundary.
Therefore, in our case, we marked all samples $x_i$, which class label probability returend by the balckbox model were below a threshold $t$ as \emph{in-danger} samples, and followed the original upsampling procedure proposed in BorderlineSMOTE~\cite{borderlinesmote2005}.
In our implementation, we used the threshold $t$ to be equal to one standard deviation away from the mean of the confidence of a blackbox model.
In other words, if we define $P_M(x_i)$ as the confidence of a model $M$ in classifying the sample $x_i$, the threshold $\delta$ is defined as follows:
$$\delta = E[P_M] - \sqrt{E[P_M^2]-E[P_M]^2}$$

Additionally, to enforce that the samples are generated only towards the directions that are the most important from the perspective of feature-importance attribution explanation mechanism such as SHAP, we generate samples by transforming existing instances by a vector defined as a negative gradient of the SHAP values.

We approximate the SHAP gradients with a model $M$ fitted to the SHAP values of a particual feature over a given dataset.
In our case we use LinearRegression as $M$, but it can be replaced by any other model gor which the gradients can be extracted easily or calcualted numerically.

$$\nabla_j = \frac{d\phi_j}{dx_j} \approx L({x}_j, \phi_j)$$

Once we obtain the gradients, we perturb each sample  towards the decision boundary using the negative gradient of the SHAP values. 
We calcualte the gradients separately for each dimensions (feature).
Let $\alpha$ be a small step size, then the perturbed value of a sample $x$ is calculated as follows:

$$x_{\text{perturbed}} = {x_j} - \alpha \cdot \nabla_j$$
Where ${x}_{\text{perturbed}} $ is a vector consisting of $x_j$ values, each perturbed in the direction defined by the gradient $\nabla_j$.

\begin{figure}
  \centering
      \includegraphics[width=1.0\columnwidth]{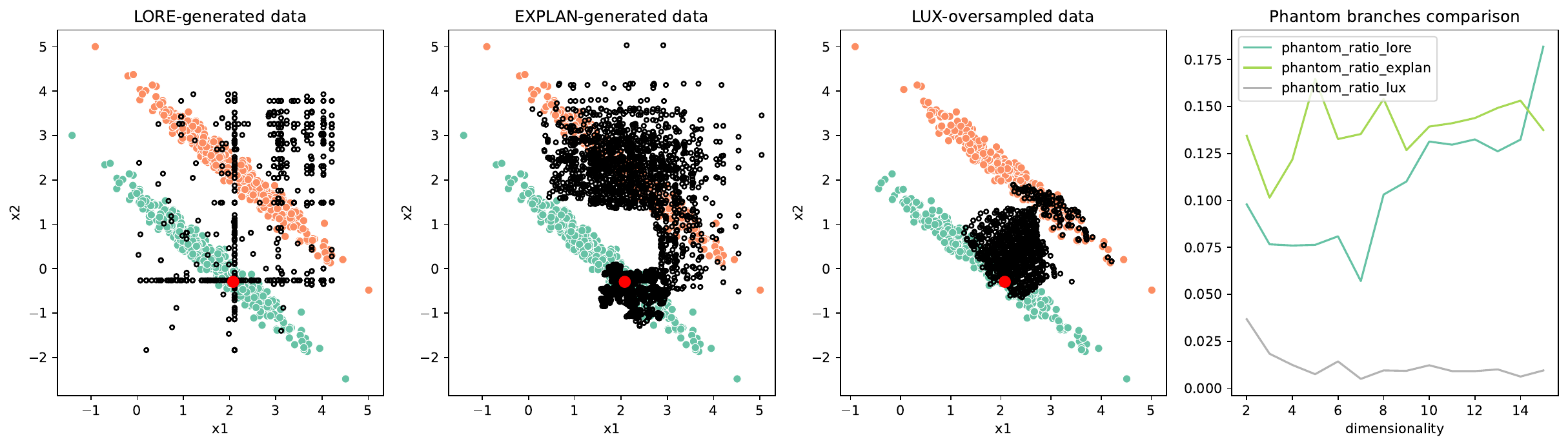}

\caption{Comparison of the LUX oversampling method with the LORE and EXPLAN approaches. The black circles represent artificially generated data. The red dot represents the instance that is being explained. The colored points represent the original data labeled by the blackbox model.}\label{fig:lux-oversampling}
\end{figure}

The motivation behind that procedure is in the theoretical justification of SHAP values, which show the impact the feature value of an instance has with respect tot he expected value of a blackbox model. 
While the expected value of a balckbox classification model for a balanced dataset is located around the decision boundary, the analysis of the gradients of SHAP values provides a convenient way of selecting a direction into which the nearest decision boundary might be located without the need of sampling large number of artificial datapoints.
The example of this mechanism is depicted in Figure~\ref{fig:lux-oversampling}, where black circles represent the replicated data in the direction of negative gradients of the SHAP values.

The example of this upsampling is presented in Figure~\ref{fig:lux-oversampling}. 
It can be seen that the upsampling does not change the shape of a distribution drastically, not affecting the phantom-branches issue discussed earlier.

The whole neighborhood selection process is governed by two main parameters: $K$, and $sigma$.
The former defines the size of the base neighborhood $N^b$, and therefore can be used to balance between local and global explanations. 
The value of $K$ equal to 1 makes the model behave like a global explainer as it will use all of the samples from the dataset, while smaller values will focuse on fraction of the neighbourhood of an instance that is being explained.
The latter defines the minimal size of density clusters that are formed to extend the $N^b \cup N_1$ neighborhood. 
The larger the parameter $\sigma$, the stronger the expansion of the base neighborhood.
Similarly as in the case of the $K$ parameter, larger values of $\sigma$ make the model generate less fine-grained local explanations.

The representative sample selection phase tries to find a balance between the local and global neighborhood to select the minimal neighborhood required to generate explanations that are influenced by high-density areas of the original data points, not local artificially generated masses.
This addresses  R1 (simplicity) because the neighborhood never gets more complicated than in the original data, and R3 (consistency) because the neighborhood selection process  is deterministic.

In the following section, we will show how we modified the tree learning algorithm in order to exploit the feature-importance explanation for obtaining simpler, more consistent and stable rule-based explantions.

\subsection{Creation of a rule-based explanation model}
The neighborhood selected according to the procedure described in previous sections is then used to build explainable rule-based model.
The model provides both factual and counterfactual explanations that are consistent (stable) with each other, because they are based on the same tree-based structure, where both factual and counterfactual explanations are single branches extracted from that tree.
For the tree generation method, we redesigned the UnertainDecisionTree classifier (UId3)~\cite{sbk2017uncertcas,sbk2018icaisc} algorithm adding the following extensions: 1) use of the feature-importance explanation model as a support method to select the optimal split, 2) incorporation of oblique splits~\cite{joao997oblique} to better capture decision boundaries of the blackbox model that are linear combinations of several attributes.
Our limits the computational cost needed of selecting features for oblique node, as we do not need to test all possible configurations, but rely on the importance obtained from SHAP values this is the main difference of this component, when compared to already existing solutions such as~\cite{obliquexai}, where all subsets of featutres are tested for the ebst split node.

We based our method on the UId3 algorithm, by extending notion of information gain split criterion.
The classic information gain formula  for the numeric feature $F$ and a training set $X$ is defined as follows:
\begin{equation}
\label{eq:infogain}
\mathit{Gain}(F,X,v) = H(X) - \left[ \frac{\|X_{<v}\|}{\|X\|} H(X_{<v}) + \frac{\|X_{\geq v}\|}{\|X\|} H(X_{\geq v}) \right]
\end{equation}
Where $X_{<v}$ is a subset of $X= \{x_i \in X : x_i<v\}$ and $H(X)$ is the entropy for the training set $X$ is defined as follows:
\begin{equation}
\label{eq:entropy}
H(X) = - \sum_{l \in Domain(F_t)} p(l) \log_{2} p(l) 
\end{equation}
Where  $Domain(F_t)$ is a  set of all classes in $X$
and $p(l)$ is a ratio of the number of elements of class label $l$ to all the elements in $X$.

We extended the Equation~(\ref{eq:infogain}) to include feature importance parameter that allows selection of split attributes that are more consistent with what the original model considers important.
It also allows one to overcome the greediness of the decision tree algorithm, because the feature importance values can be considered as the look-ahead parameters~\cite{eloma2003latrees} that favor the optimal split attributes first.
The modified information gain is given as a product of the information gain $Gain(F,X,v)$ and the importance of the feature $F_i$. 
The formula is given in Equation~(\ref{eq:luxinfogain}).
\begin{equation}
\label{eq:luxinfogain}
Gain^{LUX}(F_i,X,v) = Gain(F_i,X,v)\cdot Imp(F_i)
\end{equation}

\noindent
Where $Imp(F_i)$ is the average absolute importance of feature $F_i$  returned from feature-importance explanation mechanism such as  SHAP or LIME.

We also added the oblique splits, which allow us to create a split in decision tree that contains a condition which is a linear combination of several features.
In our approach we limited the oblique splits to use only two features because this allows us to visualize splits as 2D scatterplots.
The complete procedure for selecting an attribute and an expression to create a split in the decision tree was presented in Algorithm~\ref{alg:luxsplit}.

\begin{algorithm}[tb]
   \caption{Split selection algorithm for LUX explanation method.}
   \label{alg:luxsplit}
\begin{algorithmic}[l]
   \State {\bfseries Input:} data $X$; set of features $F$; balckbox model $M$
   \State {\bfseries Output:} split expression $E$; split attribute $F_s$
   \If{\alg{Homogeneous($X$)}}
      \State \textbf{return~}{$\varnothing$}
   \EndIf
   \State $F_i,v$ \assign {Best split attribute and value using $Gain^{LUX}(F,X,v)$}
   \State $F_{SVM}$ \assign $\left \{ F_1, F_2 \in F : max(Imp(F_1)+Imp(F_2)) \right \}$
   \State Train linear SVM classifier on $X$ and $F_{SVM}$
   \State $V_{F_1} \assign \alpha F_2 + \beta$ as a decision boundary or SVM
   \If {$Gain^{LUX}(F_i,X,v) < Gain^{LUX}(F_1,X,V_{F_1})$}
     \State $E$ \assign $\alpha F_2 + \beta$
     \State $F_s$ \assign $F_1$
   \Else
     \State $E$ \assign $v$
     \State $F_s$ \assign $F_i$
    \EndIf
    \State \textbf{return~} {split expression $E$; split attribute $F_s$}
\end{algorithmic}
\end{algorithm}

The procedure is called recursively during the tree-building algorithm.
At a given depth of a tree, the expression $E$ and the attribute $F_s$ returned by Algorithm~\ref{alg:luxsplit} are used as building blocks of a decision tree.
The $F_i$ becomes a tree node, and the expression $E$ is used as the right-hand side of a condition on a tree branch originating at node $F_i$.
The condition is denoted as $C(F)$, where $F$ is a set containing both the split feature $F_i$ and other features that can be part of expression $E$.
Finally, the explanation $\Phi^{E\rightarrow M(x_i)}$ as defined in Equation~(\ref{eq:explanation}) is extracted from a tree as a conjunction of $C(F)$ that the explained instances pass through the tree during the classification process.

\begin{figure}
  \centering
      \includegraphics[width=1.0\columnwidth]{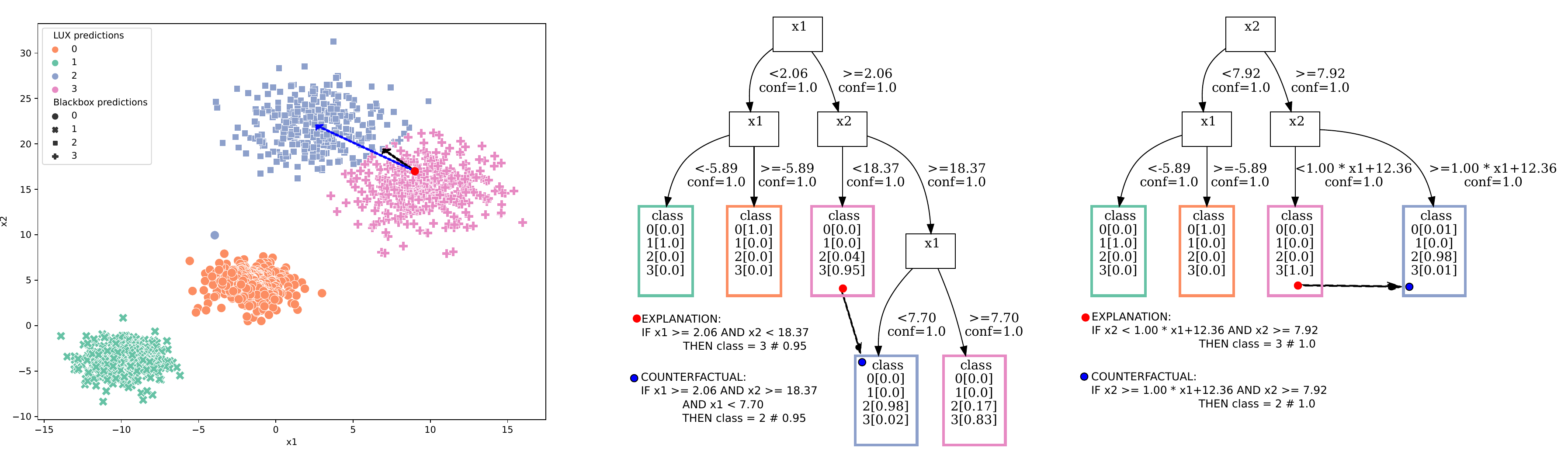}
\caption{Illustrative example of explanation generation with classic method (left tree) and shap-enhanced oblique splits we proposed. The arrow in the scatterplot represent the counterfactual explanations generated with right tree. The black arrow shows classical counterfactual example, the blue arrow shows medoid-based counterfactual example.}\label{fig:cf-lux-generation}
\end{figure}

The example of explanations generated with the LUX algorithm is given in Figure~\ref{fig:cf-lux-generation}.
The left decision tree was generated using the classical method, and the right decision tree was generated with the use of SHAP feature importances and oblique splits.
The latter is less complex (in terms of depth) and of better quality (in terms of classification accuracy).
The explanation $\Phi^{E\rightarrow M(x_i)}$ for $x_i=[9, 15]$ is created by extracting a tree branch ending in a leaf that contains the instance $x_i$.

The same tree structure is used to generate counterfactual explanations.
This ensures that the counterfactual explanation is consistent with the factual explanation in terms of the features used in a rule.
The counterfactual explanation $\Phi_{CF}^{\overline{E}\rightarrow M(x_i)}$ as defined in Equation~(\ref{eq:counterfactual}) is an explanation that tells how the given instance should change in order to be classified differently by the blackbox model.
Therefore, in a first step to generate  counterfactual explanation for an instance $x_i$, we search all branches in the explanation tree that have a leaf with majority of samples labeled differently than the $x_i$.
In the second step, we search for the nearest neighbor in instances covered by different branches.
The instance $\overline{x}_i$ that is the nearest neighbor of $x_i$ becomes a counterfactual example, and the tree branch that contains the $\overline{x}_i$ becomes the explanation $\Phi_{CF}^{\overline{E}\rightarrow M(x_i)}$.
Alternatively, to nearest neighbor one can search for the nearest medoid instead of the nearest instance.
In such a case, the $\Phi_{CF}^{\overline{E}\rightarrow M(x_i)}$ is generated based on the branch of the explanation tree that covers a set of instances which medoid is the nearest neighbor of instance $x_i$. 
In this setting, the medoid becomes the counterfactual example $\overline{x}_i$.
In Figure~\ref{fig:cf-lux-generation} both types of counterfactuals were visualized with arrows: the black represents the nearest-neighbor counterfactual, while the blue represents the medoid-based counterfactual.

Although the explanations generated with LUX are simpler (R1) and provide better quality in terms of fidelity compared with state-of-the-art methods, usability might be limited due to the difficulty in oblique splits comprehension.
In the next section, we describe how we overcome this problem by enabling simple visualizations of explanations to make them more understandable and useful.

\subsection{Explanation visualization}
\begin{figure}
  \centering
      \includegraphics[width=1.0\columnwidth]{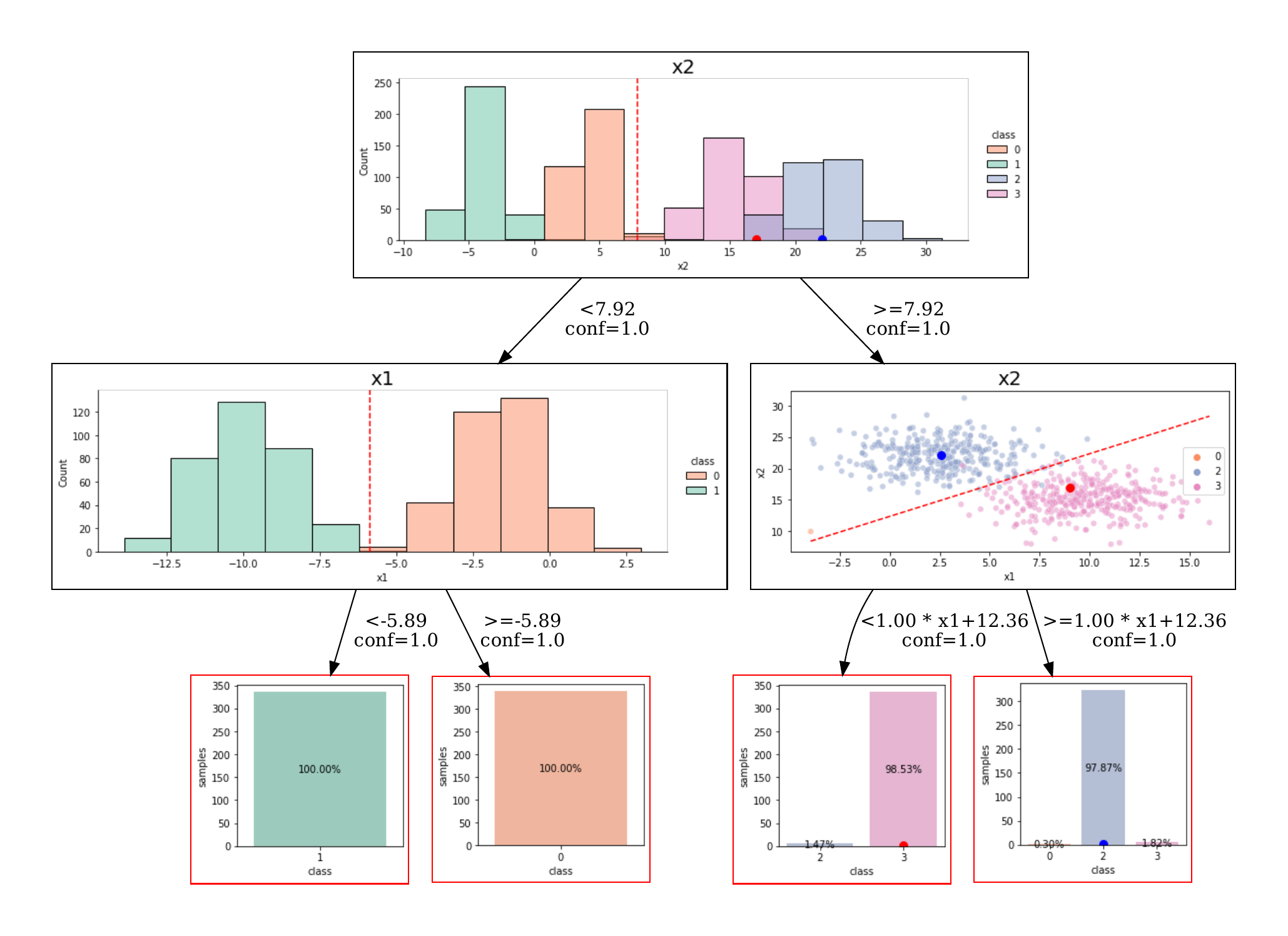}
\caption{Visual representation of an explanation tree presented in Figure~\ref{fig:cf-lux-generation}. It embeds information about data distribution at different levels of the tree and visualizes the path that explained instance (red dot) and medoid-based counterfactual passes through the tree. On the arrows, the conditions are denoted that are used to construct an explanation rule.}\label{fig:cf-tree-visual}
\end{figure}

The explanation has to be understandable by the user in order to present some value in practical applications.
In this section, we present a visualization layer that we included in our LUX explainer.
Because our explanation method is based on decision trees, we focused on visualizing this component.
The visual representation of an explanation tree from Figure~\ref{fig:cf-lux-generation} was presented in Figure~\ref{fig:cf-tree-visual}
It follows the standard structure of the tree, with nodes representing features and edges representing split conditions.
However, the nodes do not only contain the name of the feature, but also present the data distribution and a location of a split in this context (the red dotted line).
In case of oblique splits, we present the node in 2D with a linear boundary representing the split.

The visualization we propose makes the explainability process more human-readible and the explanation itself more informative.
For example, consider the explanation in a rule format, as it was presented in Figure~\ref{fig:cf-lux-generation}:
\begin{verbatim}
IF x2 < 1.00 * x1+12.36 AND x2 >= 7.92 THEN class = 3 # 1.0
\end{verbatim}
It is hardly possible to actually understand the explanation without additional context, such as the possible ranges of values for feature \verb$x2$
The visualization of this context in a form of distribution plot, with marked location of explained instance, allows the user to immediately generate counterfactuals, without any sophisticated methods.
For example, only by looking at the first split, one can notice that in order to flip decision of the model from class 3 to class 0 it is enough to change the value of \verb$x2$ to 5 and the value of \verb$x1$ to any value from a range of $(-5; 2.5)$.
Such an operation would not be possible with only the textual version of the explanation.

Furthermore, the visualization of distribution allows for expert-based quality assessment of the explanation and counterfactual.
For example, if the counterfactual explanation was located very close to the decision boundary in a scatterplot in Figure~\ref{fig:cf-tree-visual}, one can immediately judge this explanation as uncertain and even manually correct it by pushing the values of the counterfactual further away from the decision boundary.
Such a modification would not be possible with a textual explanation.

\section{Evaluation}
\label{sec:eval}

In this section, we present results from our evaluation of the LUX explainer.
We used synthetic datasets in cases where we investigated impact of the size of the dataset, and number of features on the explainability method.
In other cases, we used real, publicly available data.

Our evaluation approach was aimed at investigating the performance of LUX with respect to the previously defined four requirements R1-R4.
We compared our method with the most popular rule-based explainers discussed previously: LORE, EXPLAN and Anchor.

Our main goal was to show that we can achieve simpler, more consistent rules and counterfactuals without affecting the fidelity of the explainer.
In comparing the explainers, we used 57 real dataset and several different synthetic datasets, depending on the evaluation task.
See Table~\ref{tab:realds} for details of each of the real dataset and Table~\ref{tab:synthds} for characteristics of the synthetic datasets.

First, we collected results from all the explainers performed on the same datasets for the same instance.
Then, we performed a Friedman test followed by a Nemenyi pairwise post-hot test for multiple comparisons of mean rank sums.
This allowed us to observe how the algorithms differ, and use this unified measure co aggregated all of the metrics into one comprehensive summary.
We excluded Anchor from comparisons that involved evaluation of features that this particular explainer does not offer (e.g. counterfactuals).
The results of the evaluation are discussed below, while detailed results were presented in Appendix~\ref{sec:appB}.

For each dataset we randomly selected around 10\% of instances from test set and generated explanations for them with each of the evaluated explainers.
For instances in which the explainer could not generate any explanation, the algorithms were given minimal score.
As a measure of fidelity, and counterfactual we used the F1 score and tested it on the real datasets introduced in Table~\ref{tab:realds}.
To measure following metrics, we used eqautions introduced in Section~\ref{sec:measures}.
Due to the fact that only LORE provides the explicit way to generating counterfactual explanations, we tested this requirement by checking the number of phantom branches present in the data.

\begin{figure}[!htb]
  \centering
      \includegraphics[width=0.6\columnwidth]{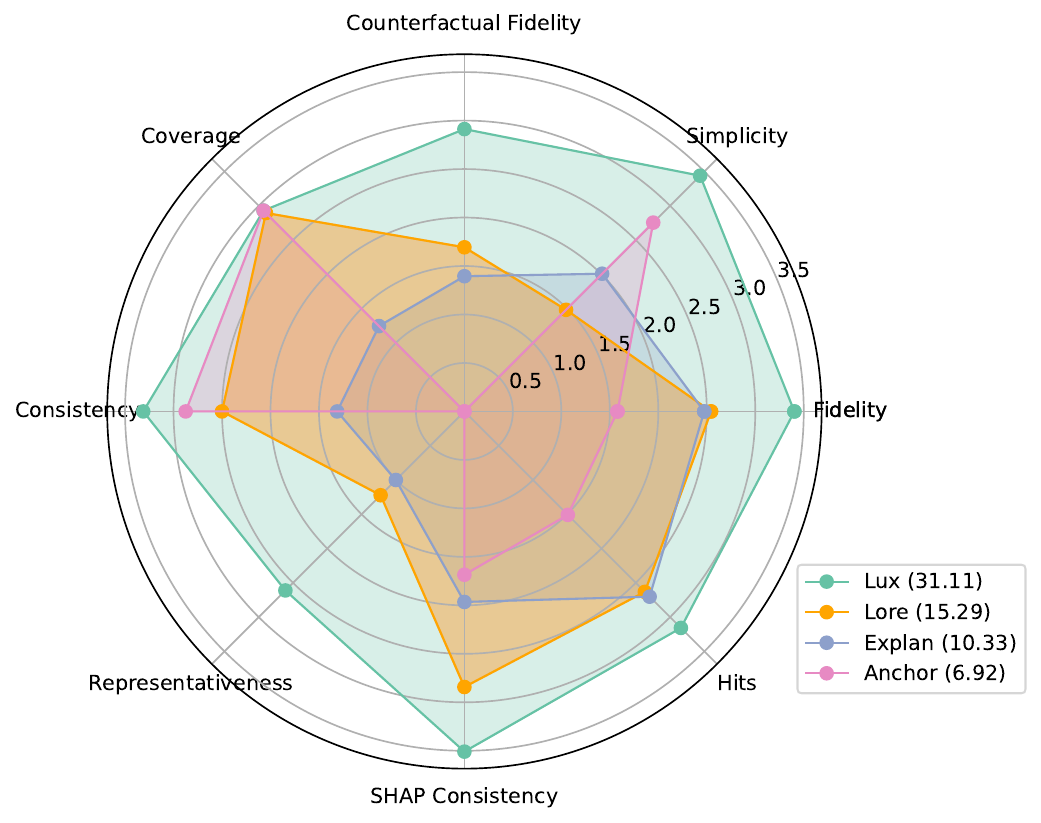}
\caption{Summary plot of all the metrics used in evaluation. The larger the area for a particular explainer, the better. The number next to the legend is the area of the polygon on the plot defined by the ranks obtained for particular methods.}\label{fig:spiderplot}
\end{figure}

In this section, we summarize the results from the evaluation and provide their concise interpretation.
The full dataset and source code used to generate results were given in supplementary materials.
The summary of these objective metrics is shown in Figure~\ref{fig:spiderplot}.
The ranks used in the figure were obtained from average performance from different tests. 
The average of the ranks that would have been assigned to all the tied values is assigned to each value.
Due to the fact that there exists trade-off between different metrics, it is desirable to maximize the area on the radar plot from Figure~\ref{fig:spiderplot}.
It can be seen that the largest area is reserved for LUX.
The tabular summarization of the plot is also given in Table~\ref{tab:evalsummary}, while detailed results for particular daatsets followed with statistical significance tests are included in Appendix~\ref{sec:appB}.
IN the summary we included two additional metrics, namely hits and coverage. 
The hits metric represent the ratio of explanations that returned correct label for the instance that is being explained.
This can be considered as point-fidelity focused on the instance that is being explained.
Coverage shows how general the explanation rule is, i.e. the more samples it covers, the more applicable it is.

\begin{table}
\caption{Average results of LUX performance for various metrics on all real datasets. The statistically significant differences were marked with $\uparrow$. }
\label{tab:evalsummary}
\scriptsize
\begin{tabularx}{\textwidth}{|p{3.5cm}|X|X|X|X|}
\hline
\textbf{Metric} & \textbf{LUX} & \textbf{EXPLAN} & \textbf{LORE} & \textbf{Anchor} \\
 \hline \hline
Consistency & \textbf{3.51 $\pm$ 0.89} $\uparrow$ & 1.96 $\pm$ 0.63 & 2.84 $\pm$ 0.70 & 1.68 $\pm$ 1.14\\ \hline
Shap Consistency & \textbf{3.51 $\pm$ 0.89} $\uparrow$& 1.96 $\pm$ 0.63 & 2.84 $\pm$ 0.70 & 1.68 $\pm$ 1.14\\ \hline
Counterfactual Fidelity & \textbf{2.91 $\pm$ 0.34} $\uparrow$& 1.39 $\pm$ 0.59 & 1.69 $\pm$ 0.50 & 0.00 $\pm$ 0.00\\ \hline
Coverage & \textbf{2.93 $\pm$ 0.96} & 1.25 $\pm$ 0.47 & 2.89 $\pm$ 0.67 & 2.93 $\pm$ 1.16\\ \hline
Fidelity & \textbf{3.40 $\pm$ 0.86} $\uparrow$& 2.47 $\pm$ 0.87 & 2.54 $\pm$ 0.95 & 1.58 $\pm$ 1.00\\ \hline
Hits & \textbf{3.16 $\pm$ 0.40} & 2.70 $\pm$ 0.68 & 2.63 $\pm$ 0.86 & 1.51 $\pm$ 0.77\\ \hline
Representativeness & \textbf{3.00 $\pm$ 0.00} $\uparrow$& 1.39 $\pm$ 0.50 & 1.61 $\pm$ 0.50 & 0.00 $\pm$ 0.00\\ \hline
Simplicity & \textbf{3.53 $\pm$ 0.87} $\uparrow$& 2.10 $\pm$ 0.74 & 1.57 $\pm$ 0.75 & 2.84 $\pm$ 1.00\\ \hline
Stability & \textbf{3.31 $\pm$ 0.85} & 1.31 $\pm$ 0.79 & 2.50 $\pm$ 0.52 & 2.88 $\pm$ 1.13\\ \hline
\end{tabularx}
\end{table}

The local fidelity is significantly higher for LUX than for trhe remaining explainers.
The counterfactuals generated with LUX exhibit noticeably higher fidelity than any other methods. 
The rules generated by LUX are the simplest in terms of the number of features used for explanation.
In addition, LUX demonstrates the highest SHAP consistency, indicating that it utilizes features that are also important according to the SHAP explainer. 
This could be attributed to the novel sampling mechanism used, which limits the number of samples and generates samples only in the directions of the largest gradients of SHAP values, pointing toward decision boundaries.
Regarding consistency, LUX stands out as the most consistent and stable explainer among all, providing similar explanations for similar instances. 

We also tested the practical aspect of the  LUX explanations.
The goal of this experiment was to check whether there is a difference between the usefulness of explanations provided in the form of rules and visual explanations and rate the overall usability of the LUX software by independant participants.
We conducted the experiment on 19 participants who had basic knowledge in the area of data science.
The participants were divided into two subgroups by random assignments: the first group had access to visualizations (12 participants) of explanations, and the other group did not (7 participants).
Both subgroups had to solve the same set of assignments related to three different instances and explanations that were generated for these instances.
The assignments included: 
\begin{enumerate}
\item (Q1) Based on the explanation provided by LUX, try to predict what will be the model output for the given instance.
\item (Q2) Change the counterfactual example that LUX provided in a way that will let you be more certain that it is correct, i.e. increase the confidence of the counterfactual being classified to the opposite class as explained instance.
\end{enumerate}

The final step of the experiment was a short survey that contained following questions:
\begin{enumerate}
\item Rate the overall experience of using LUX on a  5-point bipolar scaling method, analogous to the Likert scale.
\item Describe the biggest advantage of LUX
\item Describe the biggest disadvantage of LUX
\end{enumerate}

\begin{figure*}[htb]
    \centering
    \begin{minipage}[t]{0.52\textwidth} 
        \centering
        \includegraphics[width=\textwidth]{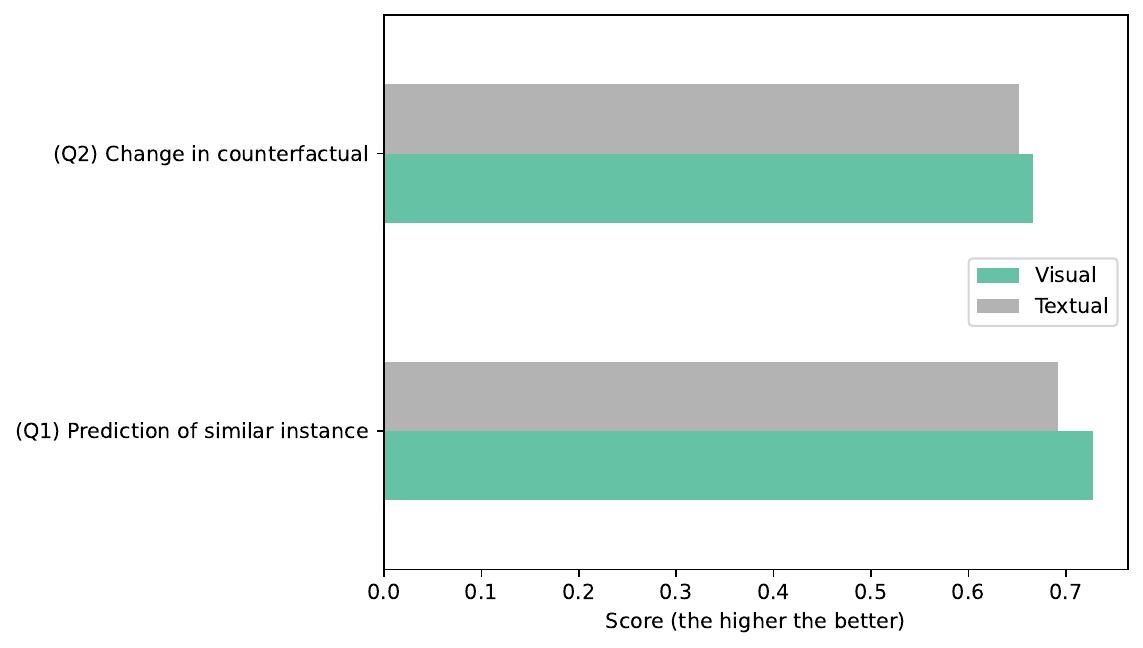}
        \vspace{0.5cm} 
    \end{minipage}\hfill
    \begin{minipage}[t]{0.45\textwidth} 
        \centering
        \includegraphics[width=\textwidth]{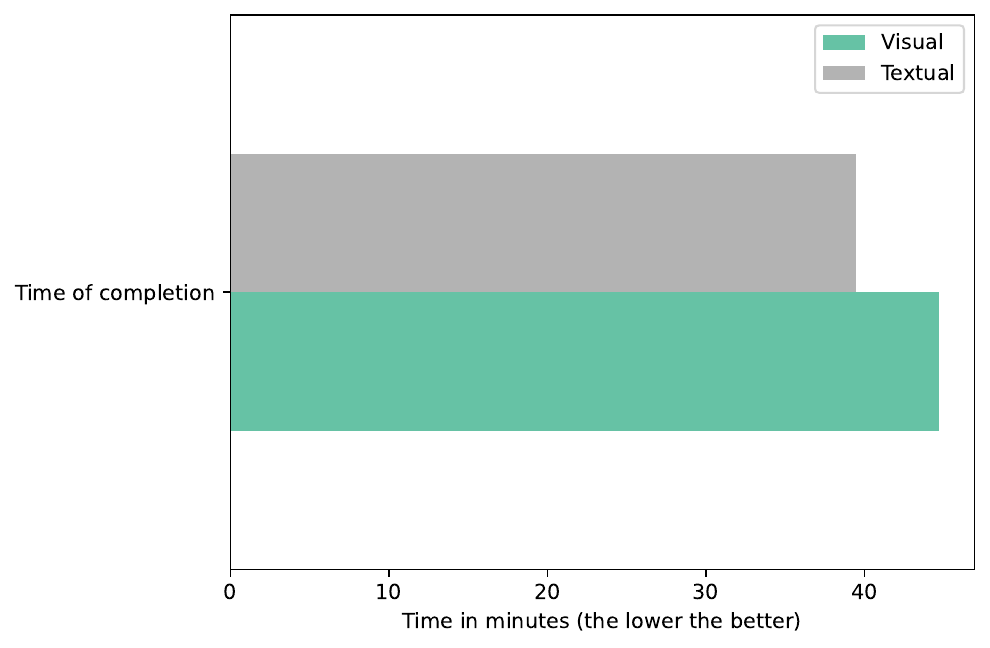}
    \end{minipage}
    \caption{(Left) Scores for assignments completed by the participants. (Right) Time of completion for all of the assignments.}
    \label{fig:combined-figures}
\end{figure*}

Additionally, we performed analysis of the assignments that the participants had to solve and measured time of completion of  this tasks.
Both groups obtained very similar results in all of the assignments and in time of completion.
The results for scores obtained for assignments (Q1) and (Q2) were visualized in left plot in Figure~\ref{fig:combined-figures}.
The scores were calculated as accuracy of human-based prediction (Q1) and accuracy of a blackbox model on a counterfactuals generated by humans (Q2).
The completion times were presented in right plot in Figure~\ref{fig:combined-figures}. 

Furthermore, both groups rated the visual and textual versions of LUX similarly, where the visual version obtained scores equal to 3.62 and the textual version 3.57.
The larger differences were observed in descriptions of advantages of  tools.
The biggest advantage pointed out by the group that had access to the visualization of the explanations was the visual aspect, which proves that it is a feature that users value.
The biggest disadvantage was the lack of detailed instructions on how to interpret the explanations.
This was a surprising finding, because the rules and decision tree formats were very well known by the participants.
It may suggest that the assumption on the simplicity of the way the knowledge is presented to the user is very subjective.
Therefore, the explanations should be more contextualized and personalized.

We observed minimal and statistically insignificant differences in responses between textual and visual versions of LUX, while comments highlighted a visual feature discrepancy.
The comments of the participants in the post-hoc survey revealed that the synthetic nature of the study presented challenges in understanding the explanations. 
This aspect carried significant weight in the final evaluation, causing participants to concentrate more on difficulties in comprehending the data and the case rather than the explanations themselves.
The primary concern was not the manner in which explanations were presented but rather how the information conveyed could have been contextualized by the intended recipients of the explanation.
The participants suggested that if the same data and explanations were provided with context, such as meaningful names for features and a description of a real-world case scenario (e.g., loan approval or income prediction), it would be much easier to understand the explanations. 
This implies the need for a greater involvement of social sciences in XAI research, particularly in investigating the informative competencies of different types of explanation recipients and formulating guidelines for personalizing XAI. 
This collaborative effort should be a mutual effort between the computer science and social science domains.

\section{Summary}
\label{sec:summary}
In this work, we presented LUX, which is a local universal rule-based explainer  that combines factual, counterfactual and visual explanations into one bundle.
It does not depend on synthetic samples, such as other state-of-the-art explainers, but rather focuses on extracting representative clusters of original data to train an interpretable decision tree classifier on top of it.
It combines feature-importance explanation methods such as SHAP to help build more stable decision trees and allows for oblique rules with can capture linear boundaries much better than any other rule-based explainer.
From the evaluation on both real and synthetic datasets, we obtained results that prove that LUX performs better in terms of simplicity, representativeness of counterfactuals and fidelity than core of the state-of-the-art explainers such as LORE, EXPLAN and Anchor.
It is also the best algorithm in terms of consistency of its explanations among consecutive runs and consistency with SHAP values.

However, it is worth keeping in mind the limitations of LUX.
One of the major ones is its computational complexity. 
For a single explanation, we need to perform three complex steps: calculation of SHAP values, neighborhood selection, and density clustering. 
It becomes very time- and memory-consuming for large datasets.
The second limitation is that LUX depends on the distance measures in the selection of neighborhood.
This works well for low-dimensional tabular data, but could be an issue in high-dimensional datasets or in other modalities such as text, images, or time series.
In such a case, one has to provide a distance metric that satisfies the different modality (text, images, time-series) or preform dimensionality reduction during neighborhood selection and clustering.

The source code of LUX,  notebooks and data used for evaluation are publicly available under the MIT license on \url{https://github.com/sbobek/lux}.

\bibliographystyle{elsarticle-num}
\bibliography{lux}

\appendix

\section{Datasets characteristics}
\label{sec:appA}
In the experiments, we used two sets of data.
The real datasets were obtained from OpenML repository in total of 57 unique datasets. 
The full characteristics of the data is presented in Table~\ref{tab:realds}.
The imbalance ratio (IR) given in the last column of the table was calculated according to the following formulae:
$\text{IR} = \frac{\max_{i=1,\ldots,K} n_i}{\min_{i=1,\ldots,K} n_i}$
where \( n_i \) denotes the number of instances in class \( i \).
The synthetic dataset was used in order to provide more controll over the data properties such as dimensionality, redundant features, number of classes and were genrated with Scikit-learn package. 
The full characteristic of the data is presented in Table~\ref{tab:synthds}.

\begin{table}
\caption{Description of the real datasets we used for evaluation.}
\label{tab:realds}
\center
\scriptsize
\begin{tabularx}{\textwidth}{|p{3.5cm}|X|X|X|}
\hline
\textbf{Dataset name} & \textbf{No of instances} & \textbf{No of features} & \textbf{Imbalance ratio}
\\ \hline \hline
GesturePhaseSegmentation & 9873.00 & 32.00 & 2.96\\ \hline
MagicTelescope & 2000.00 & 10.00 & 1.00\\ \hline
analcatdata & 841.00 & 70.00 & 5.76\\ \hline
balance-scale & 625.00 & 4.00 & 5.88\\ \hline
bank-marketing & 45211.00 & 16.00 & 7.55\\ \hline
banknote-authentication & 1372.00 & 4.00 & 1.25\\ \hline
blood-transfusion-service-center & 748.00 & 4.00 & 3.20\\ \hline
breast-w & 699.00 & 9.00 & 1.90\\ \hline
california & 2000.00 & 8.00 & 1.00\\ \hline
car & 1728.00 & 6.00 & 18.62\\ \hline
churn & 5000.00 & 20.00 & 6.07\\ \hline
climate-model-simulation-crashes & 540.00 & 18.00 & 10.74\\ \hline
cmc & 1473.00 & 9.00 & 1.89\\ \hline
covertype & 2000.00 & 10.00 & 1.00\\ \hline
credit & 2000.00 & 10.00 & 1.00\\ \hline
credit-approval & 690.00 & 15.00 & 1.25\\ \hline
credit-g & 1000.00 & 20.00 & 2.33\\ \hline
cylinder-bands & 540.00 & 37.00 & 1.37\\ \hline
diabetes & 768.00 & 8.00 & 1.87\\ \hline
disclosure & 662.00 & 3.00 & 1.09\\ \hline
dummy & 1000.00 & 6.00 & 2.66\\ \hline
electricity & 45312.00 & 8.00 & 1.36\\ \hline
eucalyptus & 736.00 & 19.00 & 2.04\\ \hline
fri & 1000.00 & 10.00 & 1.27\\ \hline
house & 2000.00 & 16.00 & 1.00\\ \hline
ilpd & 583.00 & 10.00 & 2.49\\ \hline
ilpd-numeric & 583.00 & 10.00 & 2.49\\ \hline
jungle & 44819.00 & 6.00 & 5.32\\ \hline
kc1 & 2109.00 & 21.00 & 5.47\\ \hline
kc2 & 522.00 & 21.00 & 3.88\\ \hline
kr-vs-kp & 3196.00 & 36.00 & 1.09\\ \hline
mfeat-morphological & 2000.00 & 6.00 & 1.00\\ \hline
mfeat-zernike & 2000.00 & 47.00 & 1.00\\ \hline
pc1 & 1109.00 & 21.00 & 13.40\\ \hline
pc3 & 1563.00 & 37.00 & 8.77\\ \hline
pc4 & 1458.00 & 37.00 & 7.19\\ \hline
phoneme & 5404.00 & 5.00 & 2.41\\ \hline
qsar-biodeg & 1055.00 & 41.00 & 1.96\\ \hline
rmftsa & 1024.00 & 2.00 & 4.30\\ \hline
satimage & 6430.00 & 36.00 & 2.45\\ \hline
segment & 2310.00 & 16.00 & 1.00\\ \hline
shuttle & 2000.00 & 9.00 & 786.00\\ \hline
steel-plates-fault & 1941.00 & 27.00 & 12.24\\ \hline
stock & 950.00 & 9.00 & 1.06\\ \hline
strikes & 625.00 & 6.00 & 1.02\\ \hline
texture & 5500.00 & 40.00 & 1.00\\ \hline
tic-tac-toe & 958.00 & 9.00 & 1.89\\ \hline
vehicle & 846.00 & 18.00 & 1.10\\ \hline
volcanoes-a2 & 1623.00 & 3.00 & 50.72\\ \hline
volcanoes-a3 & 1521.00 & 3.00 & 47.21\\ \hline
volcanoes-a4 & 1515.00 & 3.00 & 47.07\\ \hline
vowel & 990.00 & 12.00 & 1.00\\ \hline
wall-robot-navigation & 5456.00 & 24.00 & 6.72\\ \hline
wdbc & 569.00 & 30.00 & 1.68\\ \hline
wilt & 4839.00 & 5.00 & 17.54\\ \hline
wine & 2554.00 & 11.00 & 1.00\\ \hline
yeast & 1484.00 & 8.00 & 92.60\\ \hline
\end{tabularx}
\end{table}

\begin{table}
\caption{Description of the synthetic datasets we used for evaluation.}
\label{tab:synthds}
\center
\scriptsize
\begin{tabularx}{\textwidth}{|X|X|X|X|X|X|X|}
\hline
\textbf{Dataset name} & \textbf{No of instances} & \textbf{No of features} & \textbf{No informative features} & \textbf{No redundant features} & \textbf{No clusters per class} & \textbf{Class ratio}
\\ \hline \hline
synthetic\_1 & 1000.00 & 8.00 & 2.00 & 1.00 & 2.00 & 0.50\\ \hline
synthetic\_2 & 1250.00 & 9.00 & 2.00 & 1.00 & 2.00 & 0.50\\ \hline
synthetic\_3 & 1500.00 & 10.00 & 3.00 & 1.00 & 3.00 & 0.50\\ \hline
synthetic\_4 & 1750.00 & 11.00 & 3.00 & 1.00 & 3.00 & 0.50\\ \hline
synthetic\_5 & 2000.00 & 12.00 & 4.00 & 2.00 & 4.00 & 0.50\\ \hline
synthetic\_6 & 2250.00 & 13.00 & 4.00 & 2.00 & 4.00 & 0.50\\ \hline
synthetic\_7 & 2500.00 & 14.00 & 5.00 & 2.00 & 5.00 & 0.50\\ \hline
synthetic\_8 & 2750.00 & 15.00 & 5.00 & 2.00 & 5.00 & 0.50\\ \hline
synthetic\_9 & 3000.00 & 16.00 & 6.00 & 3.00 & 6.00 & 0.50\\ \hline
synthetic\_10 & 3250.00 & 17.00 & 6.00 & 3.00 & 6.00 & 0.50\\ \hline
synthetic\_11 & 3500.00 & 18.00 & 7.00 & 3.00 & 7.00 & 0.50\\ \hline
synthetic\_12 & 3750.00 & 19.00 & 7.00 & 3.00 & 7.00 & 0.50\\ \hline
synthetic\_13 & 4000.00 & 20.00 & 8.00 & 4.00 & 8.00 & 0.50\\ \hline
synthetic\_14 & 4250.00 & 21.00 & 8.00 & 4.00 & 8.00 & 0.50\\ \hline
synthetic\_15 & 4500.00 & 22.00 & 9.00 & 4.00 & 9.00 & 0.50\\ \hline
synthetic\_16 & 4750.00 & 23.00 & 9.00 & 4.00 & 9.00 & 0.50\\ \hline
synthetic\_17 & 5000.00 & 24.00 & 10.00 & 5.00 & 10.00 & 0.50\\ \hline
synthetic\_18 & 5250.00 & 25.00 & 10.00 & 5.00 & 10.00 & 0.50\\ \hline
synthetic\_19 & 5500.00 & 26.00 & 11.00 & 5.00 & 11.00 & 0.50\\ \hline
synthetic\_20 & 5750.00 & 27.00 & 11.00 & 5.00 & 11.00 & 0.50\\ \hline
synthetic\_21 & 6000.00 & 28.00 & 12.00 & 6.00 & 12.00 & 0.50\\ \hline
synthetic\_22 & 6250.00 & 29.00 & 12.00 & 6.00 & 12.00 & 0.50\\ \hline
synthetic\_23 & 6500.00 & 30.00 & 13.00 & 6.00 & 13.00 & 0.50\\ \hline
synthetic\_24 & 6750.00 & 31.00 & 13.00 & 6.00 & 13.00 & 0.50\\ \hline
synthetic\_25 & 7000.00 & 32.00 & 14.00 & 7.00 & 14.00 & 0.50\\ \hline
synthetic\_26 & 7250.00 & 33.00 & 14.00 & 7.00 & 14.00 & 0.50\\ \hline
synthetic\_27 & 7500.00 & 34.00 & 15.00 & 7.00 & 15.00 & 0.50\\ \hline
synthetic\_28 & 7750.00 & 35.00 & 15.00 & 7.00 & 15.00 & 0.50\\ \hline
synthetic\_29 & 8000.00 & 36.00 & 16.00 & 8.00 & 16.00 & 0.50\\ \hline
synthetic\_30 & 8250.00 & 37.00 & 16.00 & 8.00 & 16.00 & 0.50\\ \hline
synthetic\_31 & 8500.00 & 38.00 & 17.00 & 8.00 & 17.00 & 0.50\\ \hline
synthetic\_32 & 8750.00 & 39.00 & 17.00 & 8.00 & 17.00 & 0.50\\ \hline
synthetic\_33 & 9000.00 & 40.00 & 18.00 & 9.00 & 18.00 & 0.50\\ \hline
synthetic\_34 & 9250.00 & 41.00 & 18.00 & 9.00 & 18.00 & 0.50\\ \hline
synthetic\_35 & 9500.00 & 42.00 & 19.00 & 9.00 & 19.00 & 0.50\\ \hline
synthetic\_36 & 9750.00 & 43.00 & 19.00 & 9.00 & 19.00 & 0.50\\ \hline
synthetic\_37 & 10000.00 & 44.00 & 20.00 & 10.00 & 20.00 & 0.50\\ \hline
synthetic\_38 & 10250.00 & 45.00 & 20.00 & 10.00 & 20.00 & 0.50\\ \hline
synthetic\_39 & 10500.00 & 46.00 & 21.00 & 10.00 & 21.00 & 0.50\\ \hline
synthetic\_40 & 10750.00 & 47.00 & 21.00 & 10.00 & 21.00 & 0.50\\ \hline
synthetic\_41 & 11000.00 & 48.00 & 22.00 & 11.00 & 22.00 & 0.50\\ \hline
synthetic\_42 & 11250.00 & 49.00 & 22.00 & 11.00 & 22.00 & 0.50\\ \hline
synthetic\_43 & 11500.00 & 50.00 & 23.00 & 11.00 & 23.00 & 0.50\\ \hline
synthetic\_44 & 11750.00 & 51.00 & 23.00 & 11.00 & 23.00 & 0.50\\ \hline
synthetic\_45 & 12000.00 & 52.00 & 24.00 & 12.00 & 24.00 & 0.50\\ \hline
synthetic\_46 & 12250.00 & 53.00 & 24.00 & 12.00 & 24.00 & 0.50\\ \hline
\end{tabularx}
\end{table}

\section{Statistical tests}
\label{sec:appB}
For each statistical test we used Friedman test to reject null hypothesis followed by the Nemenyi post-hoc test to prove statistical significance.
We included only the tests that proves LUX is performing significantly better then other algorithms. 
In the remaining cases, although it obtains the highest scores, the difference is not statistically significant.

\subsection{Fidelity}
\begin{figure}

        \centering
      \includegraphics[width=0.6\columnwidth]{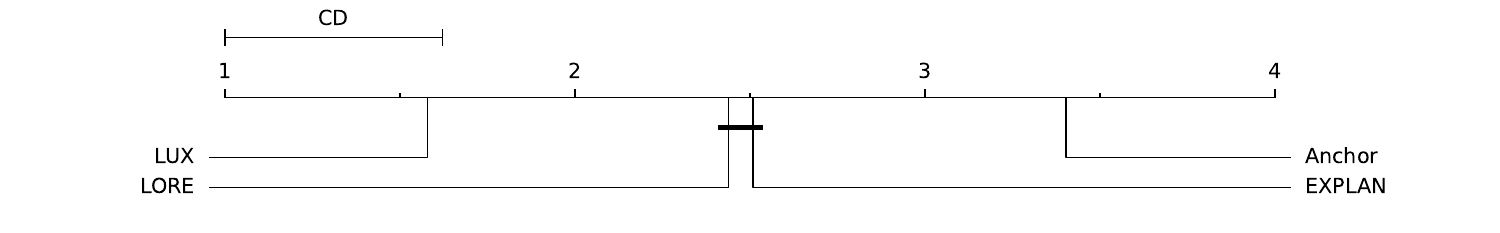}
\caption{Nemenyi post-hoc test for fidelity results from Table~\ref{tab:fidelity-results} with critical distance marked as horizontal bold line.}
\label{fig:fidelity-results}
\end{figure}

As a measure of fidelity, we used the F1 score and tested it on the real datasets introduced in Table~\ref{tab:realds}.
For instances in which the explainer could not generate any explanation, the fidelity was set to 0.
The results are presented in Table~\ref{tab:fidelity-results}.
The results of that test and the critical distance are visualized in Figure~\ref{fig:fidelity-results}.

\begin{table}
\caption{Results from the experiments on local fidelity. The value after $\pm$ corresponds to standard deviation.}
\label{tab:fidelity-results}
\scriptsize
\begin{tabularx}{\textwidth}{|p{3.5cm}|X|X|X|X|}
\hline
\textbf{Dataset name} & \textbf{LUX}&\textbf{EXPLAN}&\textbf{LORE}&\textbf{Anchor}
\\ \hline \hline
GesturePhaseSegmentation & 0.52 $\pm$ 0.38 & 0.51 $\pm$ 0.30 & 0.47 $\pm$ 0.31 & \textbf{0.55 $\pm$ 0.27}\\ \hline
MagicTelescope & \textbf{0.85 $\pm$ 0.20} & 0.81 $\pm$ 0.21 & 0.81 $\pm$ 0.19 & 0.65 $\pm$ 0.19\\ \hline
analcatdata & 0.16 $\pm$ 0.09 & \textbf{0.17 $\pm$ 0.11} & 0.16 $\pm$ 0.08 & 0.16 $\pm$ 0.28\\ \hline
balance-scale & 0.84 $\pm$ 0.28 & 0.77 $\pm$ 0.28 & \textbf{0.86 $\pm$ 0.19} & 0.73 $\pm$ 0.28\\ \hline
bank-marketing & \textbf{0.80 $\pm$ 0.20} & 0.77 $\pm$ 0.23 & 0.79 $\pm$ 0.14 & 0.65 $\pm$ 0.17\\ \hline
banknote-authentication & \textbf{1.00 $\pm$ 0.01} & 0.95 $\pm$ 0.19 & 0.99 $\pm$ 0.05 & 0.94 $\pm$ 0.14\\ \hline
blood-transfusion-service-center & 0.83 $\pm$ 0.15 & \textbf{0.84 $\pm$ 0.14} & 0.81 $\pm$ 0.15 & 0.80 $\pm$ 0.17\\ \hline
breast-w & 0.99 $\pm$ 0.10 & 0.99 $\pm$ 0.04 & \textbf{1.00 $\pm$ 0.02} & 0.85 $\pm$ 0.35\\ \hline
california & \textbf{0.87 $\pm$ 0.22} & 0.83 $\pm$ 0.21 & 0.81 $\pm$ 0.18 & 0.65 $\pm$ 0.17\\ \hline
car & \textbf{0.98 $\pm$ 0.09} & 0.94 $\pm$ 0.15 & 0.97 $\pm$ 0.10 & 0.89 $\pm$ 0.25\\ \hline
churn & \textbf{0.92 $\pm$ 0.20} & 0.90 $\pm$ 0.19 & 0.89 $\pm$ 0.21 & 0.82 $\pm$ 0.32\\ \hline
climate-model-simulation-crashes & 0.93 $\pm$ 0.17 & 0.95 $\pm$ 0.18 & 0.95 $\pm$ 0.10 & \textbf{0.98 $\pm$ 0.08}\\ \hline
cmc & \textbf{0.64 $\pm$ 0.23} & 0.55 $\pm$ 0.27 & 0.61 $\pm$ 0.25 & 0.35 $\pm$ 0.32\\ \hline
covertype & 0.75 $\pm$ 0.26 & 0.74 $\pm$ 0.24 & \textbf{0.76 $\pm$ 0.18} & 0.55 $\pm$ 0.09\\ \hline
credit & \textbf{0.78 $\pm$ 0.20} & 0.76 $\pm$ 0.22 & 0.76 $\pm$ 0.16 & 0.56 $\pm$ 0.11\\ \hline
credit-approval & 0.88 $\pm$ 0.15 & 0.86 $\pm$ 0.24 & \textbf{0.88 $\pm$ 0.15} & 0.53 $\pm$ 0.48\\ \hline
credit-g & 0.80 $\pm$ 0.25 & 0.75 $\pm$ 0.34 & \textbf{0.81 $\pm$ 0.23} & 0.59 $\pm$ 0.43\\ \hline
cylinder-bands & 0.70 $\pm$ 0.26 & \textbf{0.75 $\pm$ 0.36} & 0.75 $\pm$ 0.32 & 0.35 $\pm$ 0.44\\ \hline
diabetes & 0.72 $\pm$ 0.31 & 0.73 $\pm$ 0.32 & \textbf{0.74 $\pm$ 0.28} & 0.71 $\pm$ 0.26\\ \hline
disclosure & 0.54 $\pm$ 0.19 & 0.51 $\pm$ 0.19 & \textbf{0.54 $\pm$ 0.16} & 0.53 $\pm$ 0.11\\ \hline
dummy & \textbf{0.97 $\pm$ 0.06} & 0.94 $\pm$ 0.13 & 0.95 $\pm$ 0.08 & 0.85 $\pm$ 0.17\\ \hline
electricity & \textbf{0.80 $\pm$ 0.18} & 0.77 $\pm$ 0.20 & 0.79 $\pm$ 0.15 & 0.64 $\pm$ 0.14\\ \hline
eucalyptus & \textbf{0.61 $\pm$ 0.36} & 0.57 $\pm$ 0.40 & 0.56 $\pm$ 0.39 & 0.41 $\pm$ 0.41\\ \hline
fri & \textbf{0.86 $\pm$ 0.24} & 0.82 $\pm$ 0.23 & 0.83 $\pm$ 0.18 & 0.68 $\pm$ 0.18\\ \hline
house & \textbf{0.87 $\pm$ 0.19} & 0.84 $\pm$ 0.22 & 0.86 $\pm$ 0.18 & 0.67 $\pm$ 0.18\\ \hline
ilpd & 0.70 $\pm$ 0.31 & 0.78 $\pm$ 0.30 & 0.73 $\pm$ 0.31 & \textbf{0.79 $\pm$ 0.30}\\ \hline
ilpd-numeric & 0.75 $\pm$ 0.29 & \textbf{0.79 $\pm$ 0.26} & 0.76 $\pm$ 0.26 & 0.71 $\pm$ 0.22\\ \hline
jungle & \textbf{0.82 $\pm$ 0.22} & 0.74 $\pm$ 0.25 & 0.80 $\pm$ 0.18 & 0.58 $\pm$ 0.19\\ \hline
kc1 & \textbf{0.85 $\pm$ 0.20} & 0.80 $\pm$ 0.00 & 0.80 $\pm$ 0.00 & 0.62 $\pm$ 0.42\\ \hline
kc2 & \textbf{0.82 $\pm$ 0.22} & 0.80 $\pm$ 0.21 & 0.81 $\pm$ 0.21 & 0.56 $\pm$ 0.42\\ \hline
kr-vs-kp & \textbf{1.00 $\pm$ 0.01} & 0.97 $\pm$ 0.07 & 0.99 $\pm$ 0.03 & 0.21 $\pm$ 0.37\\ \hline
mfeat-morphological & \textbf{0.74 $\pm$ 0.26} & 0.70 $\pm$ 0.27 & 0.74 $\pm$ 0.23 & 0.42 $\pm$ 0.36\\ \hline
mfeat-zernike & 0.73 $\pm$ 0.38 & \textbf{0.77 $\pm$ 0.36} & 0.73 $\pm$ 0.32 & 0.76 $\pm$ 0.36\\ \hline
pc1 & \textbf{0.91 $\pm$ 0.11} & 0.80 $\pm$ 0.00 & 0.80 $\pm$ 0.00 & 0.87 $\pm$ 0.21\\ \hline
pc3 & \textbf{0.88 $\pm$ 0.20} & 0.80 $\pm$ 0.00 & 0.80 $\pm$ 0.00 & 0.87 $\pm$ 0.24\\ \hline
pc4 & \textbf{0.90 $\pm$ 0.21} & 0.80 $\pm$ 0.00 & 0.80 $\pm$ 0.00 & 0.30 $\pm$ 0.41\\ \hline
phoneme & \textbf{0.88 $\pm$ 0.15} & 0.84 $\pm$ 0.17 & 0.85 $\pm$ 0.17 & 0.79 $\pm$ 0.19\\ \hline
qsar-biodeg & \textbf{0.82 $\pm$ 0.28} & 0.82 $\pm$ 0.32 & 0.80 $\pm$ 0.29 & 0.75 $\pm$ 0.36\\ \hline
rmftsa & 0.84 $\pm$ 0.29 & \textbf{0.87 $\pm$ 0.28} & 0.84 $\pm$ 0.26 & 0.75 $\pm$ 0.25\\ \hline
satimage & 0.90 $\pm$ 0.16 & 0.90 $\pm$ 0.18 & \textbf{0.93 $\pm$ 0.11} & 0.89 $\pm$ 0.20\\ \hline
segment & \textbf{0.90 $\pm$ 0.17} & 0.88 $\pm$ 0.20 & 0.84 $\pm$ 0.21 & 0.71 $\pm$ 0.27\\ \hline
shuttle & \textbf{0.99 $\pm$ 0.07} & 0.80 $\pm$ 0.00 & 0.98 $\pm$ 0.09 & 0.88 $\pm$ 0.21\\ \hline
steel-plates-fault & \textbf{0.73 $\pm$ 0.31} & 0.68 $\pm$ 0.35 & 0.64 $\pm$ 0.34 & 0.65 $\pm$ 0.32\\ \hline
stock & \textbf{0.94 $\pm$ 0.18} & 0.91 $\pm$ 0.19 & 0.87 $\pm$ 0.19 & 0.79 $\pm$ 0.20\\ \hline
strikes & 0.76 $\pm$ 0.27 & \textbf{0.80 $\pm$ 0.25} & 0.74 $\pm$ 0.24 & 0.53 $\pm$ 0.14\\ \hline
texture & \textbf{0.96 $\pm$ 0.09} & 0.82 $\pm$ 0.24 & 0.81 $\pm$ 0.23 & 0.95 $\pm$ 0.08\\ \hline
tic-tac-toe & \textbf{0.85 $\pm$ 0.16} & 0.80 $\pm$ 0.25 & 0.84 $\pm$ 0.17 & 0.48 $\pm$ 0.38\\ \hline
vehicle & \textbf{0.79 $\pm$ 0.34} & 0.72 $\pm$ 0.31 & 0.76 $\pm$ 0.32 & 0.56 $\pm$ 0.26\\ \hline
volcanoes-a2 & \textbf{0.97 $\pm$ 0.10} & 0.96 $\pm$ 0.15 & 0.96 $\pm$ 0.15 & 0.96 $\pm$ 0.13\\ \hline
volcanoes-a3 & \textbf{0.93 $\pm$ 0.12} & 0.93 $\pm$ 0.16 & 0.92 $\pm$ 0.17 & 0.92 $\pm$ 0.15\\ \hline
volcanoes-a4 & \textbf{0.90 $\pm$ 0.21} & 0.89 $\pm$ 0.21 & 0.86 $\pm$ 0.25 & 0.87 $\pm$ 0.26\\ \hline
vowel & 0.84 $\pm$ 0.27 & \textbf{0.85 $\pm$ 0.25} & 0.74 $\pm$ 0.31 & 0.74 $\pm$ 0.37\\ \hline
wall-robot-navigation & 0.91 $\pm$ 0.21 & 0.78 $\pm$ 0.29 & 0.83 $\pm$ 0.19 & \textbf{0.92 $\pm$ 0.17}\\ \hline
wdbc & 0.99 $\pm$ 0.03 & 0.99 $\pm$ 0.04 & 0.98 $\pm$ 0.06 & \textbf{1.00 $\pm$ 0.02}\\ \hline
wilt & \textbf{0.98 $\pm$ 0.11} & 0.97 $\pm$ 0.10 & 0.98 $\pm$ 0.07 & 0.90 $\pm$ 0.19\\ \hline
wine & \textbf{0.78 $\pm$ 0.26} & 0.74 $\pm$ 0.24 & 0.73 $\pm$ 0.21 & 0.64 $\pm$ 0.14\\ \hline
yeast & 0.62 $\pm$ 0.24 & \textbf{0.78 $\pm$ 0.04} & 0.62 $\pm$ 0.20 & 0.50 $\pm$ 0.17\\ \hline
\end{tabularx}
\end{table}

\subsection{Simplicity}
\begin{figure}

        \centering
      \includegraphics[width=0.6\columnwidth]{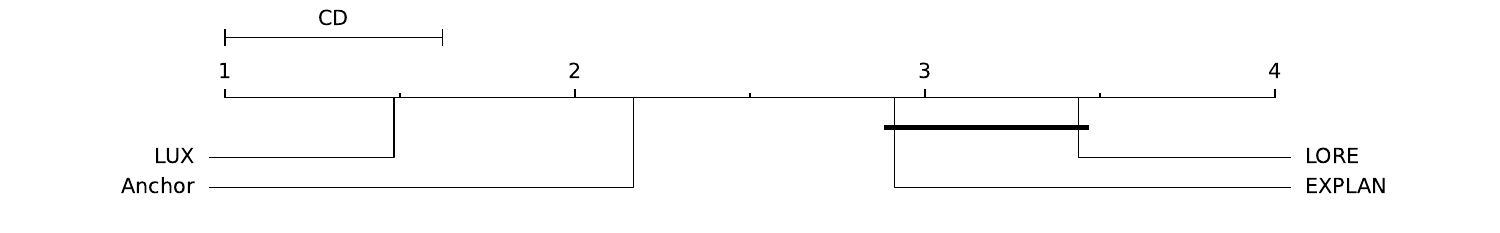}
\caption{Nemenyi post-hoc test for simplicity results from Table~\ref{tab:simplicity-results} with critical distance marked as horizontal bold line.}
\label{fig:simplicity-results}
\end{figure}

In tests for simplicity, we investigated whether we can obtain shorter rules without affecting the quality of the explainer model.
The length of the rules was calculated as the number of different attributes used in the conditions of a rule.
In this test, the smaller the value the better.
It can be seen that LUX produces significantly simpler rules explanations than other explainers, as depicted in Table~\ref{tab:simplicity-results} and Figure~\ref{fig:simplicity-results}.

From the Friedman test, we obtained statistics equal to 75.28, ith p-value equal to $0.03*10^{-13}$.
We used the same set of data sets as previously, therefore, for $\alpha=0.05$ the critical value is 2.66, which allows us to reject the null hypothesis.

\begin{table}
\caption{Results of simplicity test performed on real datasets. The lower the value, the better.}
\label{tab:simplicity-results}
\scriptsize
\begin{tabularx}{\textwidth}{|p{3.5cm}|X|X|X|X|}
\hline
\textbf{Dataset name} & \textbf{LUX}&\textbf{EXPLAN}&\textbf{LORE}&\textbf{Anchor}
\\ \hline \hline
GesturePhaseSegmentation& 0.52 $\pm$ 0.38 & 0.51 $\pm$ 0.30 & 0.47 $\pm$ 0.31 & \textbf{0.55 $\pm$ 0.27}\\ \hline
MagicTelescope & \textbf{0.85 $\pm$ 0.20} & 0.81 $\pm$ 0.21 & 0.81 $\pm$ 0.19 & 0.65 $\pm$ 0.19\\ \hline
analcatdata & 0.16 $\pm$ 0.09 & \textbf{0.17 $\pm$ 0.11} & 0.16 $\pm$ 0.08 & 0.16 $\pm$ 0.28\\ \hline
balance-scale & 0.84 $\pm$ 0.28 & 0.77 $\pm$ 0.28 & \textbf{0.86 $\pm$ 0.19} & 0.73 $\pm$ 0.28\\ \hline
bank-marketing & \textbf{0.80 $\pm$ 0.20} & 0.77 $\pm$ 0.23 & 0.79 $\pm$ 0.14 & 0.65 $\pm$ 0.17\\ \hline
banknote-authentication & \textbf{1.00 $\pm$ 0.01} & 0.95 $\pm$ 0.19 & 0.99 $\pm$ 0.05 & 0.94 $\pm$ 0.14\\ \hline
blood-transfusion-service-center & 0.83 $\pm$ 0.15 & \textbf{0.84 $\pm$ 0.14} & 0.81 $\pm$ 0.15 & 0.80 $\pm$ 0.17\\ \hline
breast-w & 0.99 $\pm$ 0.10 & 0.99 $\pm$ 0.04 & \textbf{1.00 $\pm$ 0.02} & 0.85 $\pm$ 0.35\\ \hline
california & \textbf{0.87 $\pm$ 0.22} & 0.83 $\pm$ 0.21 & 0.81 $\pm$ 0.18 & 0.65 $\pm$ 0.17\\ \hline
car & \textbf{0.98 $\pm$ 0.09} & 0.94 $\pm$ 0.15 & 0.97 $\pm$ 0.10 & 0.89 $\pm$ 0.25\\ \hline
churn & \textbf{0.92 $\pm$ 0.20} & 0.90 $\pm$ 0.19 & 0.89 $\pm$ 0.21 & 0.82 $\pm$ 0.32\\ \hline
climate-model-simulation-crashes & 0.93 $\pm$ 0.17 & 0.95 $\pm$ 0.18 & 0.95 $\pm$ 0.10 & \textbf{0.98 $\pm$ 0.08}\\ \hline
cmc & \textbf{0.64 $\pm$ 0.23} & 0.55 $\pm$ 0.27 & 0.61 $\pm$ 0.25 & 0.35 $\pm$ 0.32\\ \hline
covertype & 0.75 $\pm$ 0.26 & 0.74 $\pm$ 0.24 & \textbf{0.76 $\pm$ 0.18} & 0.55 $\pm$ 0.09\\ \hline
credit & \textbf{0.78 $\pm$ 0.20} & 0.76 $\pm$ 0.22 & 0.76 $\pm$ 0.16 & 0.56 $\pm$ 0.11\\ \hline
credit-approval & 0.88 $\pm$ 0.15 & 0.86 $\pm$ 0.24 & \textbf{0.88 $\pm$ 0.15} & 0.53 $\pm$ 0.48\\ \hline
credit-g & 0.80 $\pm$ 0.25 & 0.75 $\pm$ 0.34 & \textbf{0.81 $\pm$ 0.23} & 0.59 $\pm$ 0.43\\ \hline
cylinder-bands & 0.70 $\pm$ 0.26 & \textbf{0.75 $\pm$ 0.36} & 0.75 $\pm$ 0.32 & 0.35 $\pm$ 0.44\\ \hline
diabetes & 0.72 $\pm$ 0.31 & 0.73 $\pm$ 0.32 & \textbf{0.74 $\pm$ 0.28} & 0.71 $\pm$ 0.26\\ \hline
disclosure & 0.54 $\pm$ 0.19 & 0.51 $\pm$ 0.19 & \textbf{0.54 $\pm$ 0.16} & 0.53 $\pm$ 0.11\\ \hline
dummy & \textbf{0.97 $\pm$ 0.06} & 0.94 $\pm$ 0.13 & 0.95 $\pm$ 0.08 & 0.85 $\pm$ 0.17\\ \hline
electricity & \textbf{0.80 $\pm$ 0.18} & 0.77 $\pm$ 0.20 & 0.79 $\pm$ 0.15 & 0.64 $\pm$ 0.14\\ \hline
eucalyptus & \textbf{0.61 $\pm$ 0.36} & 0.57 $\pm$ 0.40 & 0.56 $\pm$ 0.39 & 0.41 $\pm$ 0.41\\ \hline
fri & \textbf{0.86 $\pm$ 0.24} & 0.82 $\pm$ 0.23 & 0.83 $\pm$ 0.18 & 0.68 $\pm$ 0.18\\ \hline
house & \textbf{0.87 $\pm$ 0.19} & 0.84 $\pm$ 0.22 & 0.86 $\pm$ 0.18 & 0.67 $\pm$ 0.18\\ \hline
ilpd & 0.70 $\pm$ 0.31 & 0.78 $\pm$ 0.30 & 0.73 $\pm$ 0.31 & \textbf{0.79 $\pm$ 0.30}\\ \hline
ilpd-numeric & 0.75 $\pm$ 0.29 & \textbf{0.79 $\pm$ 0.26} & 0.76 $\pm$ 0.26 & 0.71 $\pm$ 0.22\\ \hline
jungle & \textbf{0.82 $\pm$ 0.22} & 0.74 $\pm$ 0.25 & 0.80 $\pm$ 0.18 & 0.58 $\pm$ 0.19\\ \hline
kc1 & \textbf{0.85 $\pm$ 0.20} & 0.80 $\pm$ 0.00 & 0.80 $\pm$ 0.00 & 0.62 $\pm$ 0.42\\ \hline
kc2 & \textbf{0.82 $\pm$ 0.22} & 0.80 $\pm$ 0.21 & 0.81 $\pm$ 0.21 & 0.56 $\pm$ 0.42\\ \hline
kr-vs-kp & \textbf{1.00 $\pm$ 0.01} & 0.97 $\pm$ 0.07 & 0.99 $\pm$ 0.03 & 0.21 $\pm$ 0.37\\ \hline
mfeat-morphological & \textbf{0.74 $\pm$ 0.26} & 0.70 $\pm$ 0.27 & 0.74 $\pm$ 0.23 & 0.42 $\pm$ 0.36\\ \hline
mfeat-zernike & 0.73 $\pm$ 0.38 & \textbf{0.77 $\pm$ 0.36} & 0.73 $\pm$ 0.32 & 0.76 $\pm$ 0.36\\ \hline
pc1 & \textbf{0.91 $\pm$ 0.11} & 0.80 $\pm$ 0.00 & 0.80 $\pm$ 0.00 & 0.87 $\pm$ 0.21\\ \hline
pc3 & \textbf{0.88 $\pm$ 0.20} & 0.80 $\pm$ 0.00 & 0.80 $\pm$ 0.00 & 0.87 $\pm$ 0.24\\ \hline
pc4 & \textbf{0.90 $\pm$ 0.21} & 0.80 $\pm$ 0.00 & 0.80 $\pm$ 0.00 & 0.30 $\pm$ 0.41\\ \hline
phoneme & \textbf{0.88 $\pm$ 0.15} & 0.84 $\pm$ 0.17 & 0.85 $\pm$ 0.17 & 0.79 $\pm$ 0.19\\ \hline
qsar-biodeg & \textbf{0.82 $\pm$ 0.28} & 0.82 $\pm$ 0.32 & 0.80 $\pm$ 0.29 & 0.75 $\pm$ 0.36\\ \hline
rmftsa & 0.84 $\pm$ 0.29 & \textbf{0.87 $\pm$ 0.28} & 0.84 $\pm$ 0.26 & 0.75 $\pm$ 0.25\\ \hline
satimage & 0.90 $\pm$ 0.16 & 0.90 $\pm$ 0.18 & \textbf{0.93 $\pm$ 0.11} & 0.89 $\pm$ 0.20\\ \hline
segment & \textbf{0.90 $\pm$ 0.17} & 0.88 $\pm$ 0.20 & 0.84 $\pm$ 0.21 & 0.71 $\pm$ 0.27\\ \hline
shuttle & \textbf{0.99 $\pm$ 0.07} & 0.80 $\pm$ 0.00 & 0.98 $\pm$ 0.09 & 0.88 $\pm$ 0.21\\ \hline
steel-plates-fault & \textbf{0.73 $\pm$ 0.31} & 0.68 $\pm$ 0.35 & 0.64 $\pm$ 0.34 & 0.65 $\pm$ 0.32\\ \hline
stock & \textbf{0.94 $\pm$ 0.18} & 0.91 $\pm$ 0.19 & 0.87 $\pm$ 0.19 & 0.79 $\pm$ 0.20\\ \hline
strikes & 0.76 $\pm$ 0.27 & \textbf{0.80 $\pm$ 0.25} & 0.74 $\pm$ 0.24 & 0.53 $\pm$ 0.14\\ \hline
texture & \textbf{0.96 $\pm$ 0.09} & 0.82 $\pm$ 0.24 & 0.81 $\pm$ 0.23 & 0.95 $\pm$ 0.08\\ \hline
tic-tac-toe & \textbf{0.85 $\pm$ 0.16} & 0.80 $\pm$ 0.25 & 0.84 $\pm$ 0.17 & 0.48 $\pm$ 0.38\\ \hline
vehicle & \textbf{0.79 $\pm$ 0.34} & 0.72 $\pm$ 0.31 & 0.76 $\pm$ 0.32 & 0.56 $\pm$ 0.26\\ \hline
volcanoes-a2 & \textbf{0.97 $\pm$ 0.10} & 0.96 $\pm$ 0.15 & 0.96 $\pm$ 0.15 & 0.96 $\pm$ 0.13\\ \hline
volcanoes-a3 & \textbf{0.93 $\pm$ 0.12} & 0.93 $\pm$ 0.16 & 0.92 $\pm$ 0.17 & 0.92 $\pm$ 0.15\\ \hline
volcanoes-a4 & \textbf{0.90 $\pm$ 0.21} & 0.89 $\pm$ 0.21 & 0.86 $\pm$ 0.25 & 0.87 $\pm$ 0.26\\ \hline
vowel & 0.84 $\pm$ 0.27 & \textbf{0.85 $\pm$ 0.25} & 0.74 $\pm$ 0.31 & 0.74 $\pm$ 0.37\\ \hline
wall-robot-navigation & 0.91 $\pm$ 0.21 & 0.78 $\pm$ 0.29 & 0.83 $\pm$ 0.19 & \textbf{0.92 $\pm$ 0.17}\\ \hline
wdbc & 0.99 $\pm$ 0.03 & 0.99 $\pm$ 0.04 & 0.98 $\pm$ 0.06 & \textbf{1.00 $\pm$ 0.02}\\ \hline
wilt & \textbf{0.98 $\pm$ 0.11} & 0.97 $\pm$ 0.10 & 0.98 $\pm$ 0.07 & 0.90 $\pm$ 0.19\\ \hline
wine & \textbf{0.78 $\pm$ 0.26} & 0.74 $\pm$ 0.24 & 0.73 $\pm$ 0.21 & 0.64 $\pm$ 0.14\\ \hline
yeast & 0.62 $\pm$ 0.24 & \textbf{0.78 $\pm$ 0.04} & 0.62 $\pm$ 0.20 & 0.50 $\pm$ 0.17\\ \hline
\end{tabularx}
\end{table}

\subsection{SHAP Consistency}
\begin{figure}
        \centering
      \includegraphics[width=0.6\columnwidth]{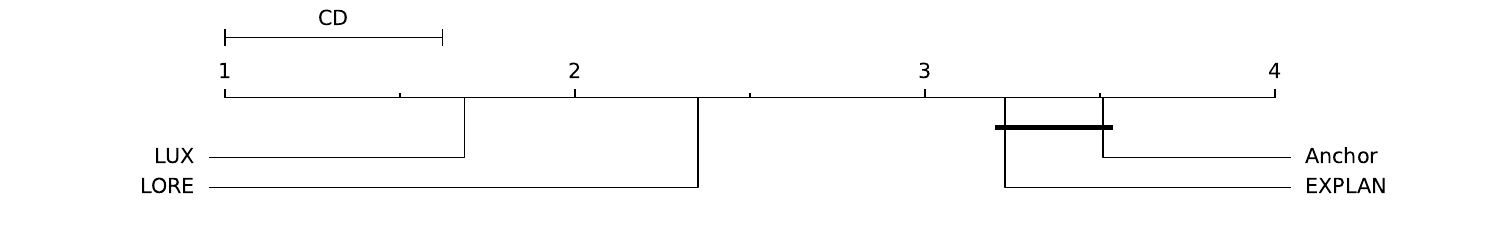}
\caption{Nemenyi post-hoc test for SHAP consistency result from Table~\ref{tab:shapconsistency} with the critical distance marked as horizontal bold line.}
\label{fig:shapconsistency-results}
\end{figure}

We tested the consistency with SHAP values of all of the explainers.
We calculated the consistency as an average SHAP importance of attributes used in rules by different explainers.
We tested this property on real datasets. 
From the Friedman test, we obtained statistics equal to 58.21, ith p-value equal to $0.023 * 10^{-10}$.
We used the same set of data sets as previously, therefore, for $\alpha=0.05$ the critical value is 2.66, which allows us to reject the null hypothesis.
The results are presented in Table~\ref{tab:shapconsistency} and visualized in Figure~\ref{fig:shapconsistency-results}.

\begin{table}
\caption{Results of consistency with SHAP test performed on real datasets. The higher the value, the better.}
\label{tab:shapconsistency}
\scriptsize
\begin{tabularx}{\textwidth}{|p{3.5cm}|X|X|X|X|}
\hline
\textbf{Dataset name} & \textbf{LUX} & \textbf{EXPLAN} & \textbf{LORE} & \textbf{Anchor}
\\ \hline \hline
GesturePhaseSegmentation & 0.04 $\pm$ 0.02 & 0.05 $\pm$ 0.02 & 0.05 $\pm$ 0.01 & \textbf{0.06 $\pm$ 0.03}\\ \hline
MagicTelescope & \textbf{0.24 $\pm$ 0.12} & 0.17 $\pm$ 0.08 & 0.18 $\pm$ 0.06 & 0.00 $\pm$ 0.00\\ \hline
analcatdata & \textbf{0.28 $\pm$ 0.03} & 0.25 $\pm$ 0.00 & 0.26 $\pm$ 0.02 & 0.21 $\pm$ 0.04\\ \hline
balance-scale & \textbf{0.29 $\pm$ 0.05} & 0.27 $\pm$ 0.04 & 0.28 $\pm$ 0.03 & 0.14 $\pm$ 0.16\\ \hline
bank-marketing & \textbf{0.36 $\pm$ 0.20} & 0.29 $\pm$ 0.17 & 0.27 $\pm$ 0.16 & 0.00 $\pm$ 0.00\\ \hline
banknote-authentication & \textbf{0.44 $\pm$ 0.16} & 0.33 $\pm$ 0.07 & 0.36 $\pm$ 0.06 & 0.19 $\pm$ 0.22\\ \hline
blood-transfusion-service-center & \textbf{0.42 $\pm$ 0.15} & 0.40 $\pm$ 0.14 & 0.41 $\pm$ 0.15 & 0.04 $\pm$ 0.13\\ \hline
breast-w & \textbf{0.25 $\pm$ 0.10} & 0.20 $\pm$ 0.08 & 0.22 $\pm$ 0.08 & 0.10 $\pm$ 0.11\\ \hline
california & \textbf{0.26 $\pm$ 0.09} & 0.23 $\pm$ 0.07 & 0.22 $\pm$ 0.07 & 0.00 $\pm$ 0.00\\ \hline
car & 0.43 $\pm$ 0.21 & 0.41 $\pm$ 0.20 & 0.45 $\pm$ 0.21 & \textbf{0.46 $\pm$ 0.21}\\ \hline
churn & 0.14 $\pm$ 0.05 & 0.10 $\pm$ 0.04 & \textbf{0.15 $\pm$ 0.08} & 0.10 $\pm$ 0.05\\ \hline
climate-model-simulation-crashes & 0.14 $\pm$ 0.09 & 0.19 $\pm$ 0.11 & 0.23 $\pm$ 0.15 & \textbf{0.25 $\pm$ 0.15}\\ \hline
cmc & \textbf{0.27 $\pm$ 0.16} & 0.19 $\pm$ 0.06 & 0.20 $\pm$ 0.12 & 0.18 $\pm$ 0.11\\ \hline
covertype & \textbf{0.34 $\pm$ 0.23} & 0.22 $\pm$ 0.12 & 0.27 $\pm$ 0.19 & 0.00 $\pm$ 0.00\\ \hline
credit & \textbf{0.32 $\pm$ 0.19} & 0.20 $\pm$ 0.10 & 0.16 $\pm$ 0.05 & 0.00 $\pm$ 0.00\\ \hline
credit-approval & \textbf{0.40 $\pm$ 0.18} & 0.19 $\pm$ 0.14 & 0.22 $\pm$ 0.11 & 0.18 $\pm$ 0.05\\ \hline
credit-g & \textbf{0.13 $\pm$ 0.05} & 0.09 $\pm$ 0.03 & 0.10 $\pm$ 0.03 & 0.10 $\pm$ 0.05\\ \hline
cylinder-bands & 0.08 $\pm$ 0.05 & 0.05 $\pm$ 0.02 & 0.08 $\pm$ 0.07 & \textbf{0.09 $\pm$ 0.07}\\ \hline
diabetes & \textbf{0.29 $\pm$ 0.13} & 0.21 $\pm$ 0.09 & 0.25 $\pm$ 0.12 & 0.12 $\pm$ 0.13\\ \hline
disclosure & \textbf{0.44 $\pm$ 0.14} & 0.42 $\pm$ 0.14 & 0.41 $\pm$ 0.11 & 0.00 $\pm$ 0.00\\ \hline
dummy & \textbf{0.25 $\pm$ 0.05} & 0.22 $\pm$ 0.05 & 0.25 $\pm$ 0.04 & 0.00 $\pm$ 0.00\\ \hline
electricity & \textbf{0.39 $\pm$ 0.22} & 0.32 $\pm$ 0.18 & 0.33 $\pm$ 0.19 & 0.00 $\pm$ 0.00\\ \hline
eucalyptus & \textbf{0.17 $\pm$ 0.10} & 0.10 $\pm$ 0.03 & 0.10 $\pm$ 0.05 & 0.11 $\pm$ 0.05\\ \hline
fri & \textbf{0.28 $\pm$ 0.11} & 0.22 $\pm$ 0.09 & 0.24 $\pm$ 0.08 & 0.00 $\pm$ 0.00\\ \hline
house & \textbf{0.17 $\pm$ 0.08} & 0.11 $\pm$ 0.05 & 0.12 $\pm$ 0.06 & 0.00 $\pm$ 0.00\\ \hline
ilpd & \textbf{0.20 $\pm$ 0.12} & 0.15 $\pm$ 0.09 & 0.18 $\pm$ 0.11 & 0.20 $\pm$ 0.11\\ \hline
ilpd-numeric & 0.16 $\pm$ 0.09 & 0.17 $\pm$ 0.10 & \textbf{0.17 $\pm$ 0.08} & 0.00 $\pm$ 0.00\\ \hline
jungle & 0.23 $\pm$ 0.05 & 0.19 $\pm$ 0.05 & \textbf{0.23 $\pm$ 0.06} & 0.00 $\pm$ 0.00\\ \hline
kc1 & 0.11 $\pm$ 0.06 & 0.20 $\pm$ 0.00 & \textbf{0.22 $\pm$ 0.00} & 0.05 $\pm$ 0.08\\ \hline
kc2 & 0.10 $\pm$ 0.08 & 0.08 $\pm$ 0.02 & 0.10 $\pm$ 0.03 & \textbf{0.11 $\pm$ 0.05}\\ \hline
kr-vs-kp & \textbf{0.28 $\pm$ 0.15} & 0.21 $\pm$ 0.16 & 0.21 $\pm$ 0.14 & 0.10 $\pm$ 0.06\\ \hline
mfeat-morphological & \textbf{0.20 $\pm$ 0.03} & 0.18 $\pm$ 0.02 & 0.19 $\pm$ 0.02 & 0.09 $\pm$ 0.09\\ \hline
mfeat-zernike & \textbf{0.05 $\pm$ 0.02} & 0.03 $\pm$ 0.01 & 0.04 $\pm$ 0.02 & 0.03 $\pm$ 0.02\\ \hline
pc1 & 0.08 $\pm$ 0.15 & 0.20 $\pm$ 0.00 & \textbf{0.22 $\pm$ 0.00} & 0.04 $\pm$ 0.05\\ \hline
pc3 & 0.08 $\pm$ 0.06 & 0.20 $\pm$ 0.00 & \textbf{0.22 $\pm$ 0.00} & 0.07 $\pm$ 0.04\\ \hline
pc4 & 0.08 $\pm$ 0.08 & 0.20 $\pm$ 0.00 & \textbf{0.22 $\pm$ 0.00} & 0.08 $\pm$ 0.05\\ \hline
phoneme & \textbf{0.32 $\pm$ 0.11} & 0.26 $\pm$ 0.11 & 0.29 $\pm$ 0.10 & 0.04 $\pm$ 0.09\\ \hline
qsar-biodeg & \textbf{0.09 $\pm$ 0.05} & 0.05 $\pm$ 0.02 & 0.06 $\pm$ 0.03 & 0.08 $\pm$ 0.06\\ \hline
rmftsa & \textbf{0.39 $\pm$ 0.13} & 0.25 $\pm$ 0.10 & 0.29 $\pm$ 0.11 & 0.00 $\pm$ 0.00\\ \hline
satimage & \textbf{0.04 $\pm$ 0.02} & 0.03 $\pm$ 0.01 & 0.04 $\pm$ 0.02 & 0.03 $\pm$ 0.02\\ \hline
segment & \textbf{0.13 $\pm$ 0.06} & 0.10 $\pm$ 0.02 & 0.11 $\pm$ 0.03 & 0.02 $\pm$ 0.04\\ \hline
shuttle & \textbf{0.23 $\pm$ 0.07} & 0.20 $\pm$ 0.00 & 0.21 $\pm$ 0.05 & 0.00 $\pm$ 0.00\\ \hline
steel-plates-fault & 0.06 $\pm$ 0.02 & 0.05 $\pm$ 0.02 & 0.06 $\pm$ 0.02 & \textbf{0.07 $\pm$ 0.03}\\ \hline
stock & \textbf{0.21 $\pm$ 0.08} & 0.16 $\pm$ 0.05 & 0.18 $\pm$ 0.05 & 0.00 $\pm$ 0.00\\ \hline
strikes & \textbf{0.41 $\pm$ 0.19} & 0.30 $\pm$ 0.15 & 0.31 $\pm$ 0.14 & 0.00 $\pm$ 0.00\\ \hline
texture & \textbf{0.04 $\pm$ 0.02} & 0.03 $\pm$ 0.02 & 0.03 $\pm$ 0.01 & 0.03 $\pm$ 0.02\\ \hline
tic-tac-toe & \textbf{0.20 $\pm$ 0.05} & 0.18 $\pm$ 0.08 & 0.18 $\pm$ 0.04 & 0.19 $\pm$ 0.10\\ \hline
vehicle & \textbf{0.10 $\pm$ 0.04} & 0.07 $\pm$ 0.02 & 0.08 $\pm$ 0.02 & 0.01 $\pm$ 0.03\\ \hline
volcanoes-a2 & 0.75 $\pm$ 0.16 & 0.58 $\pm$ 0.22 & \textbf{0.77 $\pm$ 0.14} & 0.00 $\pm$ 0.00\\ \hline
volcanoes-a3 & \textbf{0.72 $\pm$ 0.18} & 0.54 $\pm$ 0.20 & 0.70 $\pm$ 0.21 & 0.00 $\pm$ 0.00\\ \hline
volcanoes-a4 & \textbf{0.66 $\pm$ 0.21} & 0.52 $\pm$ 0.19 & 0.64 $\pm$ 0.20 & 0.00 $\pm$ 0.00\\ \hline
vowel & \textbf{0.13 $\pm$ 0.05} & 0.11 $\pm$ 0.01 & 0.11 $\pm$ 0.02 & 0.12 $\pm$ 0.04\\ \hline
wall-robot-navigation & 0.11 $\pm$ 0.10 & 0.08 $\pm$ 0.03 & 0.10 $\pm$ 0.05 & \textbf{0.12 $\pm$ 0.06}\\ \hline
wdbc & 0.05 $\pm$ 0.03 & 0.04 $\pm$ 0.02 & 0.06 $\pm$ 0.03 & \textbf{0.07 $\pm$ 0.03}\\ \hline
wilt & \textbf{0.57 $\pm$ 0.21} & 0.37 $\pm$ 0.09 & 0.45 $\pm$ 0.11 & 0.09 $\pm$ 0.24\\ \hline
wine & \textbf{0.15 $\pm$ 0.05} & 0.12 $\pm$ 0.04 & 0.14 $\pm$ 0.04 & 0.00 $\pm$ 0.00\\ \hline
yeast & \textbf{0.17 $\pm$ 0.04} & 0.12 $\pm$ 0.02 & 0.15 $\pm$ 0.05 & 0.00 $\pm$ 0.00\\ \hline
\end{tabularx}
\end{table}

\subsection{Consistency}
We tested the consistency of the explanations by calculating the Jaccard index for the set of features used in consecutive calls of explanation for the same instance.
The different configurations of datasets were selected to observe the relation between the number of features, size of the dataset, and size of noise (non-infromative features) to the stability of explanations.

\begin{figure}

        \centering
      \includegraphics[width=0.6\columnwidth]{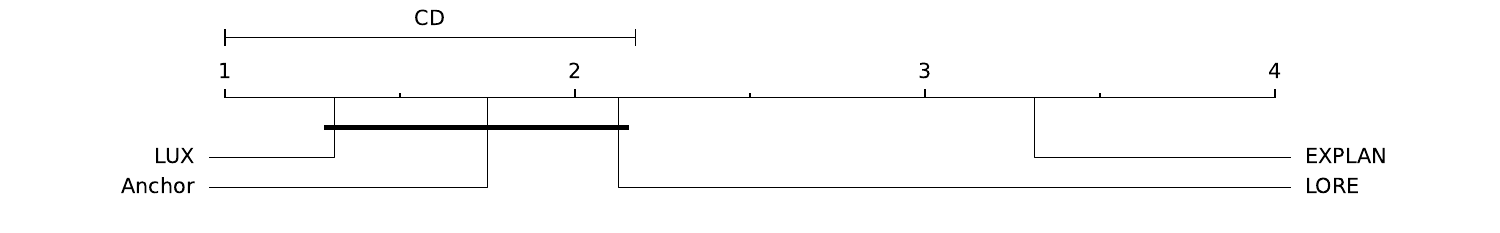}
\caption{Nemenyi post-hoc test for consistency results from Table~\ref{tab:synthds-consistency} with critical distance marked as horizontal bold line.}
\label{fig:consistency-results}
\end{figure}

From the Friedman test, we obtained statistics equal to 21.49, ith p-value equal to $0.831*10^{-4}$.
With 4 algorithms and 46 datasets we had 3 x 135 degree of freedom respectively, which allowed us to determine that the critical value for $F(3,135)$ for $\alpha=0.05$ is 2.67.
The results are presented in Table~\ref{tab:synthds-consistency} and visualized in Figure~\ref{fig:consistency-results}.

\begin{table}
\caption{Results of consistency test performed on synthetic datasets. The higher the value, the better.}
\label{tab:synthds-consistency}
\scriptsize
\begin{tabularx}{\textwidth}{|p{2.5cm}|X|X|X|X|}
\hline
\textbf{Dimensionality} & \textbf{LUX}&\textbf{EXPLAN}&\textbf{LORE}&\textbf{Anchor}
\\ \hline \hline
4 & 0.79 & 0.76 & 0.84 & 1.00\\ \hline
5 & 0.75 & 1.00 & 0.79 & 0.65\\ \hline
6 & 1.00 & 0.79 & 0.91 & 0.75\\ \hline
7 & 1.00 & 0.71 & 0.76 & 1.00\\ \hline
8 & 1.00 & 0.49 & 0.94 & 1.00\\ \hline
9 & 0.95 & 0.52 & 0.63 & 0.77\\ \hline
10 & 0.62 & 0.48 & 0.76 & 1.00\\ \hline
11 & 0.84 & 0.52 & 0.79 & 0.58\\ \hline
12 & 0.89 & 0.54 & 0.88 & 0.48\\ \hline
13 & 0.80 & 0.46 & 0.87 & 0.94\\ \hline
14 & 0.81 & 0.44 & 0.67 & 0.50\\ \hline
15 & 0.91 & 0.31 & 0.62 & 0.76\\ \hline
16 & 0.83 & 0.39 & 0.63 & 0.67\\ \hline
17 & 0.80 & 0.52 & 0.66 & 0.94\\ \hline
18 & 0.86 & 0.45 & 0.69 & 0.91\\ \hline
19 & 0.95 & 0.54 & 0.69 & 0.80\\ \hline
20 & 0.89 & 0.40 & 0.55 & 0.68\\ \hline
21 & 0.80 & 0.32 & 0.62 & 0.76\\ \hline
22 & 0.74 & 0.36 & 0.60 & 0.82\\ \hline
23 & 0.93 & 0.30 & 0.54 & 0.60\\ \hline
24 & 0.71 & 0.31 & 0.50 & 0.69\\ \hline
25 & 0.83 & 0.30 & 0.55 & 0.60\\ \hline
26 & 0.93 & 0.29 & 0.53 & 0.42\\ \hline
27 & 1.00 & 0.37 & 0.61 & 0.81\\ \hline
28 & 0.96 & 0.26 & 0.42 & 0.42\\ \hline
29 & 0.70 & 0.28 & 0.55 & 0.48\\ \hline
30 & 0.88 & 0.29 & 0.43 & 0.63\\ \hline
31 & 0.84 & 0.24 & 0.45 & 0.36\\ \hline
32 & 0.99 & 0.30 & 0.37 & 0.55\\ \hline
33 & 0.66 & 0.25 & 0.37 & 0.63\\ \hline
34 & 0.93 & 0.29 & 0.43 & 0.55\\ \hline
35 & 0.91 & 0.24 & 0.50 & 0.39\\ \hline
36 & 0.42 & 0.27 & 0.45 & 0.66\\ \hline
37 & 0.93 & 0.25 & 0.52 & 0.63\\ \hline
38 & 0.86 & 0.19 & 0.40 & 0.57\\ \hline
39 & 0.96 & 0.23 & 0.43 & 0.44\\ \hline
40 & 0.73 & 0.23 & 0.44 & 0.48\\ \hline
41 & 0.76 & 0.20 & 0.39 & 0.85\\ \hline
42 & 0.88 & 0.22 & 0.42 & 0.31\\ \hline
43 & 0.86 & 0.26 & 0.39 & 0.61\\ \hline
44 & 0.74 & 0.24 & 0.64 & 0.62\\ \hline
45 & 0.92 & 0.21 & 0.42 & 0.32\\ \hline
46 & 0.97 & 0.21 & 0.41 & 0.42\\ \hline
47 & 0.95 & 0.20 & 0.36 & 0.41\\ \hline
48 & 0.92 & 0.22 & 0.40 & 0.59\\ \hline
49 & 0.93 & 0.15 & 0.40 & 0.42\\ \hline
\end{tabularx}
\end{table}

\subsection{Counterfactuals and representativeness}
Due to the fact that only LORE provides the explicit way to generating counterfactual explanations, we tested this requirement by checking the number of phantom branches present in the data.
Additionally, we used global fidelity of the explanation trees of LORE, EXPLAN and LUX to define average fidelity of all possible counterfactuals. 
To measure the global fidelity, we used the F1 score, as in case of local fidelity.
We excluded Anchor as a method which was not designed to generate counterfactuals.

\begin{figure}

        \centering
      \includegraphics[width=0.6\columnwidth]{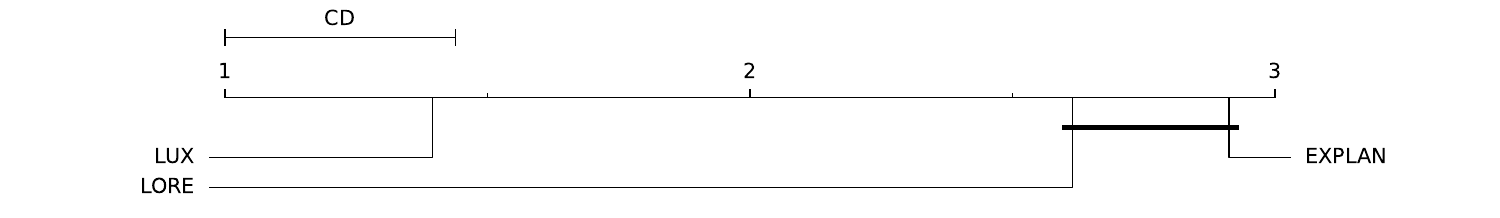}
\caption{Nemenyi post-hoc test for counterfactual fidelity result from Table~\ref{tab:cf-fidelity} with the critical distance marked as horizontal bold line.}
\label{fig:cf-fidelity}
\end{figure}

With 3 algorithms and 57 datasets we had 2 x 112 degree of freedom respectively, which allowed us to determine that the critical value for $F(2,112)$ for $\alpha=0.05$ is 3.08.
From the Friedman test, we obtained statistics equal to 74.01, ith p-value equal to $0.028 *10^{-10}$.
From the results presented in Table~\ref{tab:cf-fidelity} and the statistical tests presented in Figure~\ref{fig:cf-fidelity}.

\begin{table}
\caption{Counterfactual fidelity of all possible counterfactuals.}
\label{tab:cf-fidelity}
\scriptsize
\begin{tabularx}{\textwidth}{|p{3.5cm}|X|X|X|}
\hline
\textbf{Dataset name} &\textbf{ LUX}&\textbf{EXPLAN}&\textbf{LORE}
\\ \hline \hline
GesturePhaseSegmentation & \textbf{0.76 $\pm$ 0.17} & 0.54 $\pm$ 0.23 & 0.50 $\pm$ 0.29\\ \hline
MagicTelescope & \textbf{0.89 $\pm$ 0.08} & 0.79 $\pm$ 0.15 & 0.79 $\pm$ 0.16\\ \hline
analcatdata & 0.76 $\pm$ 0.18 & \textbf{0.81 $\pm$ 0.11} & 0.80 $\pm$ 0.14\\ \hline
balance-scale & \textbf{0.81 $\pm$ 0.15} & 0.70 $\pm$ 0.13 & 0.78 $\pm$ 0.17\\ \hline
bank-marketing & \textbf{0.92 $\pm$ 0.07} & 0.89 $\pm$ 0.11 & 0.87 $\pm$ 0.12\\ \hline
banknote-authentication & \textbf{0.99 $\pm$ 0.02} & 0.91 $\pm$ 0.14 & 0.95 $\pm$ 0.08\\ \hline
blood-transfusion-service-center & \textbf{0.97 $\pm$ 0.06} & 0.92 $\pm$ 0.11 & 0.95 $\pm$ 0.09\\ \hline
breast-w & \textbf{0.98 $\pm$ 0.05} & 0.98 $\pm$ 0.05 & 0.98 $\pm$ 0.06\\ \hline
california & \textbf{0.91 $\pm$ 0.05} & 0.84 $\pm$ 0.14 & 0.78 $\pm$ 0.15\\ \hline
car & \textbf{0.91 $\pm$ 0.10} & 0.83 $\pm$ 0.11 & 0.89 $\pm$ 0.10\\ \hline
churn & \textbf{0.96 $\pm$ 0.09} & 0.85 $\pm$ 0.19 & 0.92 $\pm$ 0.11\\ \hline
climate-model-simulation-crashes & \textbf{0.94 $\pm$ 0.10} & 0.86 $\pm$ 0.11 & 0.91 $\pm$ 0.12\\ \hline
cmc & \textbf{0.84 $\pm$ 0.08} & 0.77 $\pm$ 0.10 & 0.77 $\pm$ 0.09\\ \hline
covertype & \textbf{0.90 $\pm$ 0.06} & 0.82 $\pm$ 0.11 & 0.82 $\pm$ 0.10\\ \hline
credit & \textbf{0.93 $\pm$ 0.06} & 0.87 $\pm$ 0.10 & 0.83 $\pm$ 0.12\\ \hline
credit-approval & 0.98 $\pm$ 0.04 & 0.94 $\pm$ 0.10 & \textbf{0.98 $\pm$ 0.06}\\ \hline
credit-g & \textbf{0.85 $\pm$ 0.13} & 0.75 $\pm$ 0.15 & 0.83 $\pm$ 0.12\\ \hline
cylinder-bands & \textbf{0.84 $\pm$ 0.12} & 0.81 $\pm$ 0.14 & 0.82 $\pm$ 0.18\\ \hline
diabetes & 0.87 $\pm$ 0.14 & \textbf{0.87 $\pm$ 0.12} & 0.85 $\pm$ 0.16\\ \hline
disclosure & \textbf{0.92 $\pm$ 0.10} & 0.89 $\pm$ 0.12 & 0.87 $\pm$ 0.14\\ \hline
dummy & \textbf{0.92 $\pm$ 0.09} & 0.89 $\pm$ 0.07 & 0.91 $\pm$ 0.08\\ \hline
electricity & \textbf{0.93 $\pm$ 0.07} & 0.87 $\pm$ 0.11 & 0.87 $\pm$ 0.11\\ \hline
eucalyptus & \textbf{0.68 $\pm$ 0.14} & 0.55 $\pm$ 0.17 & 0.55 $\pm$ 0.17\\ \hline
fri & \textbf{0.88 $\pm$ 0.09} & 0.83 $\pm$ 0.13 & 0.83 $\pm$ 0.14\\ \hline
house & \textbf{0.88 $\pm$ 0.07} & 0.82 $\pm$ 0.12 & 0.76 $\pm$ 0.17\\ \hline
ilpd & \textbf{0.91 $\pm$ 0.09} & 0.83 $\pm$ 0.16 & 0.82 $\pm$ 0.18\\ \hline
ilpd-numeric & \textbf{0.88 $\pm$ 0.12} & 0.82 $\pm$ 0.16 & 0.81 $\pm$ 0.18\\ \hline
jungle & \textbf{0.80 $\pm$ 0.09} & 0.67 $\pm$ 0.11 & 0.74 $\pm$ 0.10\\ \hline
kc1 & \textbf{0.94 $\pm$ 0.16} & 0.80 $\pm$ 0.00 & 0.81 $\pm$ 0.00\\ \hline
kc2 & \textbf{0.96 $\pm$ 0.13} & 0.94 $\pm$ 0.12 & 0.95 $\pm$ 0.13\\ \hline
kr-vs-kp & \textbf{0.99 $\pm$ 0.02} & 0.92 $\pm$ 0.11 & 0.92 $\pm$ 0.08\\ \hline
mfeat-morphological & \textbf{0.86 $\pm$ 0.14} & 0.68 $\pm$ 0.26 & 0.76 $\pm$ 0.22\\ \hline
mfeat-zernike & \textbf{0.69 $\pm$ 0.13} & 0.46 $\pm$ 0.13 & 0.52 $\pm$ 0.19\\ \hline
pc1 & \textbf{0.99 $\pm$ 0.07} & 0.80 $\pm$ 0.00 & 0.81 $\pm$ 0.00\\ \hline
pc3 & \textbf{0.97 $\pm$ 0.12} & 0.80 $\pm$ 0.00 & 0.81 $\pm$ 0.00\\ \hline
pc4 & \textbf{0.96 $\pm$ 0.10} & 0.80 $\pm$ 0.00 & 0.81 $\pm$ 0.00\\ \hline
phoneme & \textbf{0.94 $\pm$ 0.08} & 0.89 $\pm$ 0.13 & 0.89 $\pm$ 0.14\\ \hline
qsar-biodeg & \textbf{0.81 $\pm$ 0.09} & 0.70 $\pm$ 0.15 & 0.66 $\pm$ 0.21\\ \hline
rmftsa & \textbf{0.92 $\pm$ 0.10} & 0.90 $\pm$ 0.13 & 0.92 $\pm$ 0.11\\ \hline
satimage & \textbf{0.89 $\pm$ 0.13} & 0.66 $\pm$ 0.24 & 0.71 $\pm$ 0.25\\ \hline
segment & \textbf{0.90 $\pm$ 0.11} & 0.74 $\pm$ 0.21 & 0.74 $\pm$ 0.21\\ \hline
shuttle & \textbf{0.99 $\pm$ 0.04} & 0.80 $\pm$ 0.00 & 0.96 $\pm$ 0.09\\ \hline
steel-plates-fault & \textbf{0.78 $\pm$ 0.16} & 0.61 $\pm$ 0.23 & 0.66 $\pm$ 0.23\\ \hline
stock & \textbf{0.95 $\pm$ 0.09} & 0.84 $\pm$ 0.21 & 0.88 $\pm$ 0.14\\ \hline
strikes & \textbf{0.86 $\pm$ 0.10} & 0.84 $\pm$ 0.15 & 0.85 $\pm$ 0.11\\ \hline
texture & \textbf{0.86 $\pm$ 0.09} & 0.50 $\pm$ 0.21 & 0.55 $\pm$ 0.20\\ \hline
tic-tac-toe & \textbf{0.85 $\pm$ 0.08} & 0.80 $\pm$ 0.13 & 0.83 $\pm$ 0.10\\ \hline
vehicle & \textbf{0.74 $\pm$ 0.14} & 0.55 $\pm$ 0.19 & 0.59 $\pm$ 0.18\\ \hline
volcanoes-a2 & \textbf{0.99 $\pm$ 0.03} & 0.97 $\pm$ 0.07 & 0.99 $\pm$ 0.04\\ \hline
volcanoes-a3 & \textbf{0.99 $\pm$ 0.06} & 0.97 $\pm$ 0.08 & 0.98 $\pm$ 0.05\\ \hline
volcanoes-a4 & \textbf{0.97 $\pm$ 0.09} & 0.92 $\pm$ 0.13 & 0.95 $\pm$ 0.11\\ \hline
vowel & \textbf{0.66 $\pm$ 0.10} & 0.43 $\pm$ 0.10 & 0.44 $\pm$ 0.13\\ \hline
wall-robot-navigation & \textbf{0.91 $\pm$ 0.04} & 0.64 $\pm$ 0.10 & 0.68 $\pm$ 0.12\\ \hline
wdbc & \textbf{0.96 $\pm$ 0.05} & 0.94 $\pm$ 0.08 & 0.94 $\pm$ 0.09\\ \hline
wilt & \textbf{0.98 $\pm$ 0.06} & 0.93 $\pm$ 0.09 & 0.92 $\pm$ 0.12\\ \hline
wine & \textbf{0.87 $\pm$ 0.08} & 0.79 $\pm$ 0.13 & 0.76 $\pm$ 0.16\\ \hline
yeast & 0.85 $\pm$ 0.11 & \textbf{0.90 $\pm$ 0.06} & 0.82 $\pm$ 0.11\\ \hline
\end{tabularx}
\end{table}

We also measured the representativeness of counterfactual as a fraction of branches in a decision tree that has zero-coverage in test set (phantom branches).
We used for that purpose the synthetic datasets, where we increase the dimensionality and measure the fraction of phantom branches.
The results are given in Table~\ref{tab:phantoms}, where it can be clearly seen that only in the case of LUX the fraction of phantom branches does not increase with the dimensionality and remains at the reasonably low level.
Other methods which rely on generating synthetic samples suffer from that issue, which may result in producing counterfactual rules that are impossible to achieve in reality.

\begin{figure}

        \centering
      \includegraphics[width=0.6\columnwidth]{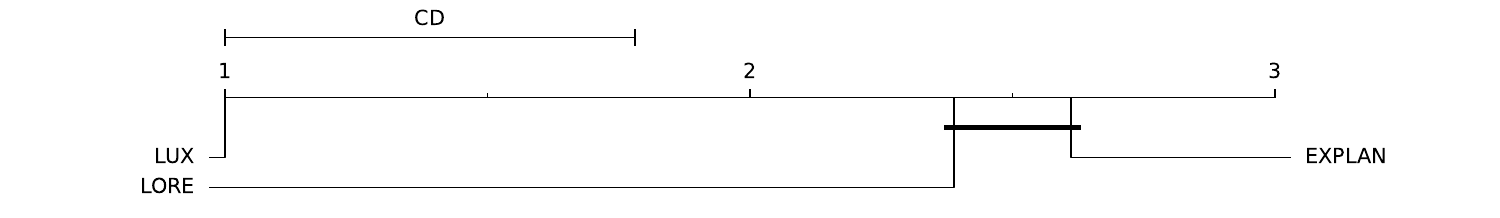}
\caption{Nemenyi post-hoc test for representativeness from Table~\ref{tab:phantoms} results with the critical distance marked as horizontal bold line.}
\label{fig:cf-representativeness}
\end{figure}

From the Friedman test, we obtained statistics equal to 27.44, ith p-value equal to $0.07*10*{-15}$.
With 3 algorithms and 14 datasets we had 2 x 90 degree of freedom respectively, which allowed us to determine that the critical value for $F(2,90)$ for $\alpha=0.05$ is 3.09.
The results from  statistical tests presented in Figure~\ref{fig:cf-representativeness}.

\begin{table}
\caption{Results representing fraction of phantom-branches in a explanation tree built by different explainers.}
\label{tab:phantoms}
\scriptsize
\begin{tabularx}{\textwidth}{|X|X|X|X|}
\hline
\textbf{Dimensionality} &\textbf{ LUX}&\textbf{EXPLAN}&\textbf{LORE}
\\ \hline \hline
4 & 0.04 & 0.15 & 0.10\\ \hline
5 & 0.01 & 0.16 & 0.07\\ \hline
6 & 0.00 & 0.21 & 0.07\\ \hline
7 & 0.00 & 0.15 & 0.11\\ \hline
8 & 0.00 & 0.18 & 0.12\\ \hline
9 & 0.02 & 0.21 & 0.10\\ \hline
10 & 0.01 & 0.17 & 0.26\\ \hline
11 & 0.01 & 0.19 & 0.15\\ \hline
12 & 0.00 & 0.17 & 0.18\\ \hline
13 & 0.00 & 0.18 & 0.14\\ \hline
14 & 0.00 & 0.12 & 0.17\\ \hline
15 & 0.00 & 0.15 & 0.15\\ \hline
16 & 0.00 & 0.14 & 0.21\\ \hline
17 & 0.00 & 0.21 & 0.15\\ \hline
18 & 0.00 & 0.19 & 0.14\\ \hline
19 & 0.00 & 0.15 & 0.17\\ \hline
20 & 0.00 & 0.17 & 0.24\\ \hline
21 & 0.00 & 0.17 & 0.14\\ \hline
22 & 0.00 & 0.17 & 0.24\\ \hline
23 & 0.00 & 0.19 & 0.22\\ \hline
24 & 0.00 & 0.16 & 0.24\\ \hline
25 & 0.00 & 0.16 & 0.23\\ \hline
26 & 0.00 & 0.16 & 0.18\\ \hline
27 & 0.00 & 0.17 & 0.25\\ \hline
28 & 0.00 & 0.17 & 0.24\\ \hline
29 & 0.00 & 0.16 & 0.28\\ \hline
30 & 0.00 & 0.14 & 0.20\\ \hline
31 & 0.00 & 0.17 & 0.27\\ \hline
32 & 0.00 & 0.17 & 0.29\\ \hline
33 & 0.00 & 0.12 & 0.28\\ \hline
34 & 0.00 & 0.14 & 0.21\\ \hline
35 & 0.00 & 0.21 & 0.27\\ \hline
36 & 0.00 & 0.15 & 0.32\\ \hline
37 & 0.00 & 0.16 & 0.24\\ \hline
38 & 0.00 & 0.16 & 0.27\\ \hline
39 & 0.00 & 0.18 & 0.22\\ \hline
40 & 0.00 & 0.17 & 0.28\\ \hline
41 & 0.00 & 0.17 & 0.26\\ \hline
42 & 0.00 & 0.16 & 0.27\\ \hline
43 & 0.00 & 0.16 & 0.26\\ \hline
44 & 0.00 & 0.17 & 0.23\\ \hline
45 & 0.00 & 0.17 & 0.30\\ \hline
46 & 0.00 & 0.17 & 0.26\\ \hline
47 & 0.00 & 0.16 & 0.30\\ \hline
48 & 0.00 & 0.14 & 0.27\\ \hline
49 & 0.00 & 0.19 & 0.28\\ \hline
\end{tabularx}
\end{table}

\end{document}